\documentclass[Afour,sageh,times,doublespace]{sagej}

\usepackage{moreverb,url}

\usepackage{graphics}
\usepackage{comment}
\usepackage[latin1]{inputenc}
\usepackage{xspace}
\usepackage{subfigure}
\usepackage{amsmath}
\usepackage{url}
\usepackage{color}
\usepackage{amsmath, amsfonts, amssymb}
\usepackage{xspace}
\usepackage{natbib}
\usepackage[ruled,lined]{algorithm2e}
\usepackage{tabulary}
\usepackage{listings}
\usepackage{blindtext}
\usepackage[normalem]{ulem}
\usepackage{censor}
\usepackage{etoolbox}
\makeatletter
\patchcmd\@combinedblfloats{\box\@outputbox}{\unvbox\@outputbox}{}{%
	\errmessage{\noexpand\@combinedblfloats could not be patched}%
}%

\newlength\nextcharwidth
\makeatletter
\renewcommand\@cenword[1]{%
	\setlength{\nextcharwidth}{\widthof{#1}}%
	\censorrule{\nextcharwidth}%
	\kern -\nextcharwidth%
	#1}
\makeatother

\bibpunct{(}{)}{;}{a}{,}{,}
\renewcommand{\cite}{\citep}
\newcommand{\rs}{\textsc{RoboSherlock}}
\newcommand{\rsshort}{\textit{perceive}}

\lstdefinelanguage[OWL]{XML} 
{morekeywords={Individual,ObjectProperty,Types,Facts,Class,SubClassOf,Domain,
Range,SubPropertyOf,EquivalentTo}}

\lstdefinestyle{OWL}    {language=[OWL]XML, lineskip=-0.3ex, fontadjust=true, 
basicstyle={\scriptsize\nopagebreak[4]}}

\lstdefinestyle{Prolog} {language=Prolog,   lineskip=-0.3ex, fontadjust=true, 
basicstyle={\small\nopagebreak[4]}, commentstyle=\scriptsize, 
showstringspaces=false, showspaces=false, showtabs=false}

\newcommand{\knowrob}{\textsc{KnowRob}\xspace}
\newcommand{\robosherlock}{{\textsc{RoboSherlock}\xspace}}
\newcommand{\myem}[1]{\textbf{\textit{#1}}}

\newcommand{\sofa}{\textsc{SofA}}

\newcommand{\strike}[1]{\textcolor{red}{\sout{}}}
\newcommand\soutref[1]{\censorruledepth=.55ex\censor{}}
\newcommand{\newtext}[1]{\textcolor{black}{#1}}

\theoremstyle{definition}
\newtheorem{mydef}{Definition}

\graphicspath{{.}}

\begin{document}

\title{\textsc{RoboSherlock}: Cognition-enabled Robot Perception for Everyday Manipulation Tasks}

\runninghead{B\'alint-Bencz\'edi et al.}

\author{Ferenc B\'alint-Bencz\'edi\affilnum{1}, Jan-Hendrik Worch\affilnum{1}, Daniel Nyga\affilnum{1}, Nico Blodow\affilnum{2}, Patrick Mania\affilnum{1}, Zolt\'an-Csaba M\'arton\affilnum{3} and Michael Beetz\affilnum{1}}

\affiliation{\affilnum{1}Institute for Artificial Intelligence, University of Bremen, DE\\
			\affilnum{2}Intelligent Autnomous Systems Group, Technical University of Munich, DE (currently with Fyuse Inc.)\\
			\affilnum{3}Robotics and Mechatronics Institute, German Aerospace Center (DLR)} 

\corrauth{Ferenc B\'alint-Bencz\'edi, 
	Institute for Artificial Intelligence,
	University of Bremen,
	Bremen
	28359 DE.}

\email{balintbe@cs.uni-bremen.de}

\begin{abstract}
A pressing question when designing intelligent autonomous systems is how to integrate the various subsystems concerned with complementary tasks. More specifically, robotic vision must provide task-relevant information about the environment and the objects in it to various planning related modules. In most implementations of the traditional Perception-Cognition-Action paradigm these tasks are treated as quasi-independent modules that function as black boxes for each other. It is our view 	that perception can benefit tremendously from a tight collaboration with cognition.	
We present \robosherlock\, a knowledge-enabled cognitive perception systems for mobile robots performing human-scale everyday manipulation tasks. In \robosherlock\, perception and interpretation of realistic scenes is formulated as an unstructured information management(UIM) problem. The application of the UIM principle supports the implementation of perception systems that can answer task-relevant queries about objects in a scene, boost object recognition performance by combining the strengths of multiple perception algorithms, support knowledge-enabled reasoning about objects and enable automatic and knowledge-driven generation of processing pipelines. We demonstrate the potential of the proposed framework through feasibility studies of systems for real-world scene perception that have been built on top of the framework.
\end{abstract}

\keywords{robot perception, knowledge-enabled perception, service robotics, RGB-D perception}

\maketitle


\section{Introduction}
\label{sec:intro}

\begin{figure}[t]
	\centering
	\includegraphics[width = 0.99\columnwidth]{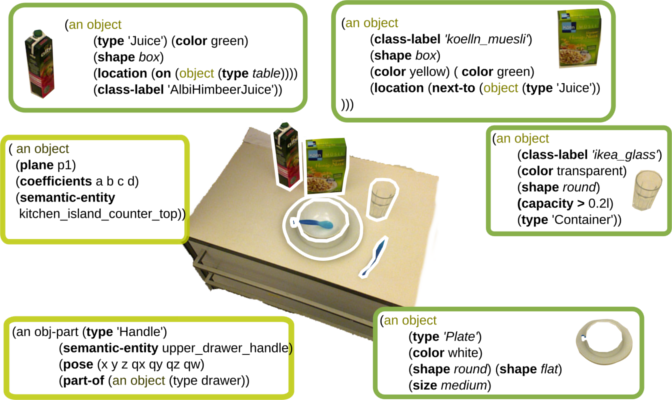}
	\caption{Semantically rich description of a breakfast scene}
	\label{fig:example_scene}
\end{figure}

\newtext{The primary role of perception in autonomous robots performing everyday 
manipulation is to extract the information needed to accomplish a respective 
task. For example, if a robot gets the task to fill a cup with milk, the container with the milk needs to be found. This might require finding the milk container in the cluttered fridge. If there are multiple milk containers the intended one has to be identified --- the open one or the one with the skim milk instead of the soy milk. The milk might be in different kinds of containers, such as bottles, milk cartons, or a bowl. It might also be necessary to recognize functional parts of the container such as the handle or the mouth of the container. In addition, the weight of the container, the center of mass, or whether the container is slippery, stable, or fragile might be important. In order to pour the milk into the cup the opening of the cup should at the top and the mouth of the container positioned above the center of cup. More generally speaking, objects often have to be perceived in difficult scenarios, or are visually challenging. Even during a simple task, such as setting the table, the variation of visual characteristics of objects is high. They can be either textured or textureless, shiny or opaque, can be occluded or can have functional parts that might be important in order to manipulate them. When it comes to manipulating an object, knowing it's label (i.e. recognizing the type of the object) is not sufficient. It is important to know the state of the object, its function, how and where to grasp it etc. 
As a second example consider the breakfast scene shown in Figure~\ref{fig:example_scene}. Given the task to clean the table and place objects at their appropriate locations, there are several challenges the perception system of a robot needs to overcome. First the objects need to be found and identified. Algorithmic approaches that might work for some (e.g. table-top segmentation of 3D clusters or key-point based CAD model fitting for the juice and cereal) will fail for others (the transparent glass, or the small flat knife). In order for a robot to safely remove the plate, the bowl has to identified on top of it and the fact that it is not empty detected. To optimize a robots plan detecting objects that are semantically similar and need to be placed in the same cupboard or drawer should be grouped (i.e. the cutlery goes in to the drawer, dirty items go in the dishwasher etc.). To place objects into these containers they first need to be opened, thus the handle of the drawer or cupboard needs to be found. Robust algorithmic solutions exist for all of these problems, but no single algorithm can cope with all the complexities of such scenarios. To enable a robot to competently \textit{perceive everything}, the need arises for a coherent combination of all the different approaches.
Obtaining each of these information pieces can be considered a separate perception task. Therefore, the perception system of an autonomous robot performing human-level manipulation actions must be capable of accomplishing a large variety of perception tasks.}

Today's robot perception systems typically only provide a small subset of the required perception functionality. At the same time most robot perception systems provide their functionality in an overly general manner. They try to accomplish the perception task by running 
predefined detection, categorization, identification, and interpretation 
algorithms mostly on images as their sole input. This means that the perception routines cannot exploit prior knowledge, the structure of manipulation tasks and the environment, and that the robot cannot prepare for future perception tasks. There is a mismatch between the methods applied to interpret the raw data and the semantic structure of the underlying information. 
For example, overly generalized system can't exploit that most people know that in a cupboard, plates are stacked and therefore appear as horizontal lines in the images at their known location. If plates have a texture most system don't know that the pattern cannot be used after a meal because the plates might have become dirty.
The environment and its structure are also not exploited to their fullest, i.e. in slowly changing environments many perception tasks or sub-tasks thereof can be performed earlier in order to be prepared when the actual task is issued.
 
These observations suggest that the perception tasks that are currently formulated and performed by robotic agents are both too easy and too hard. They are too easy because they only cover a subset of the needed functionality under context conditions. These conditions often cannot be met by applications. Therefore, they cannot enable open and competent manipulation in realistic environments. They are thus too hard because they are intractable once context conditions are not met and regularities prior knowledge has to offer in order to simplify perception tasks are not exploited. \newtext{As a simple example of this seemingly contradictory nature of perception tasks consider the following: it is easy to create a specific handle detector for the scene in Figure~\ref{fig:example_scene}, but this will only work under the specific condition that the robot is located in this exact (or very similar) environment. Creating a general handle detector that will work in all possible conditions for any kind of drawer handle is an intractable task. Hence representing and exploiting the knowledge about the environment is essential.}
\strike{In this article we propose \robosherlock, an open source software
framework for implementing perception systems for robots performing human-scale everyday manipulation tasks that facilitate the realization of perception systems with needed functionality and specialized to leverage existing structure and knowledge.} 

\newtext{In this article we propose \robosherlock, an open source software framework for robots performing human-scale everyday manipulation tasks. The framework facilitates the realization of perception systems with the needed functionality to specialize and leverage existing structure and knowledge.} \robosherlock\ enables programmers to combine perception, representation, and reasoning methods in order to scale the perception capabilities of robots towards the needs implied by general manipulation tasks. To this end \robosherlock\ supports the implementation of perception systems that
\begin{itemize}
\item[\dots] \myem{can be tasked}. The robot control program can
request the perception system to detect objects that satisfy a given
description, ask it to examine aspects of the detected objects such
as their 3D form, pose, state, etc. In this sense, perception can be
viewed as a question answering system that answers the queries of
the robot's control system based on perceived scenes.
\item[\dots] \myem{can be equipped with ensembles of expert perception
    algorithms} with complementary, similar or overlapping
  functionality instead of relying on one particular perception
  algorithm. \robosherlock\ provides control mechanisms and data
  structures to direct the algorithms to synergistically cooperate, to communicate relevant information, fuse their results and hence to combine the strengths of individual methods.
\item[\dots] \myem{can enhance perception with knowledge and reasoning.} In \robosherlock, the robot can reason about the objects to be detected and the respective task and environment context to make the perceptual processes faster, more efficient and robust. Knowledge processing also helps the system to interpret the results returned by the perception algorithms and thereby increase the set of perceptual tasks that can be accomplished. It also enables the robot to specialize perception tasks it has to solve in order to resolve ambiguities in the perception data.
\end{itemize}

 \strike{\robosherlock\ has been designed with two major implementational aspects in 
mind: (1) it does not replace any existing perception system or algorithm but 
rather enables easy integration of previous work in a unifying framework that 
allows these systems to synergistically work together and (2) new methods can 
be easily wrapped into \robosherlock\ processing modules to extend and improve 
existing functionality and performance. In order to provide these services and scale towards realistic sets of objects, environments, and perceptual tasks, \robosherlock\ considers perception as \emph{content analytics in unstructured data}, first introduced by}\soutref{~\citet{blodow14phd}}\strike{. Content Analytics (CA) denotes 
the discipline of applying methods from the field of statistical data analysis 
to large amounts of data in order to extract semantically meaningful knowledge 
from those. The data are considered \emph{unstructured} since they lack 
explicit semantics and structure. The paradigm of \emph {unstructured 
information management} (UIM) offers an implementational framework for 
realizing high-performance CA systems. Perhaps the most prominent example of a 
UIM system is \emph {Watson}}\soutref{~\cite{FerrucciEtAl10aimag}}\strike{, a question answering 
system that has won the US quiz show \emph{jeopardy!}, competing against the 
champions of the show and demonstrating an unprecedented breadth of knowledge.  
In UIM, pieces of unstructured data, such as web pages, text documents or 
images are processed by a collection of specialized information extraction 
algorithms (\emph{annotators}), and each algorithm contributes pieces of 
knowledge with respect to its expertise. Thereby, outputs of different 
algorithms are allowed to be complementary, overlapping or even contradictory. 
Hence subsequently, the collected annotations are rated and consolidated to 
come to a consistent final decision.}
\newtext{It is our belief that these three characteristics are essential if we want to scale the perceptual capabilities towards complex manipulation tasks. The ``taskability`` by the high level control system can be addressed through the means of treating perception as a question-answering problem and considering individual perceptual challenges in the context of the task at hand. Using a generic query-language, valuable background knowledge can be asserted into the perception framework that can help simplify the perception task. Having a standardized system where building blocks can be adapted based on a description, addresses the core problem of scaling perception to a large variety of tasks.}

The work presented in this article is based on two of our previous works~\cite{beetz15robosherlock, balintbe16task} and extends these with further specifications, implementation details and experiments. We focus on three main topics: first, treating perception as a question answering process on 
images and thereby boosting the \newtext{adaptability of perception systems to the given tasks}. \strike{robustness, efficiency, and accuracy of perception in open and realistic conditions showcased by perception for autonomous robots.} Second, equipping perception systems with the capability of specializing perception processes in order to exploit regularities and prior knowledge to dynamically simplify perception tasks and thereby leverage methods that are tailored for the respective tasks and contexts.
Third, interpret perception results through knowledge-informed reasoning. Additionally we also present a formal definition of parts of the system, in an attempt to standardize future contributions to the framework.

\newtext{The rest of the paper is structured as follows. In the next chapter we will offer a short introduction to what we understand under perception for everyday manipulation followed by a high level overview of the framework in Section~\ref{sec:uima_perc}. 
The conceptual description of the framework is given in Section ~\ref{sec:perc_as_uim}, formally describing all major components of the framework. We introduce and detail a perception system implemented using \robosherlock in Section ~\ref{sec:implementation}.  In Section~\ref{sec:experiments} we present three different robotic tasks where \robosherlock~was used as the main perception engine. We conclude the paper with a survey of related literature (Section ~\ref{sec:related_work}) and discussions (Section~\ref{sec:discussion}).}


\section{Perception for Everyday Manipulation tasks}
\label{sec:overview}


Everyday robot manipulation tasks usually take a considerable amount of 
time to execute, are in some sense repetitive in their nature and 
require interaction with the environment. Examples of such tasks are 
setting or cleaning a table, performing chemical experiments, or 
monitoring the stock levels of products in a supermarket. \newtext{As 
\citet{beetz15robosherlock} argue, in order to accomplish those tasks in 
human environments proficiently, object perception is highly 
task-dependent. Depending on the current context and the action a robot 
is supposed to conduct, different perceptional cues and functional 
parts of objects become important. In the case when the robot is to put 
a cooking pot on the stove, for instance, it must detect its handles in 
order to grasp it. If it is to pour water into it, however, the handles 
are less relevant but rather the center of its top opening matters. As 
it is infeasible to exhaustively process a robots sensor readings for 
all possible aspects of the world all the time, its perception 
component must be able to answer targeted questions about the 
environment on demand. In addition, for the world being only partially 
observable in nearly all real-world scenarios, it is inevitable for a 
robot to maintain an internal representation of which state the robot 
believes the environment to be in. We call this representation the 
\emph{belief state} or \emph{world model} of the robot.}

\begin{figure}[t]
    \centering
    \includegraphics[width = 0.99\columnwidth]{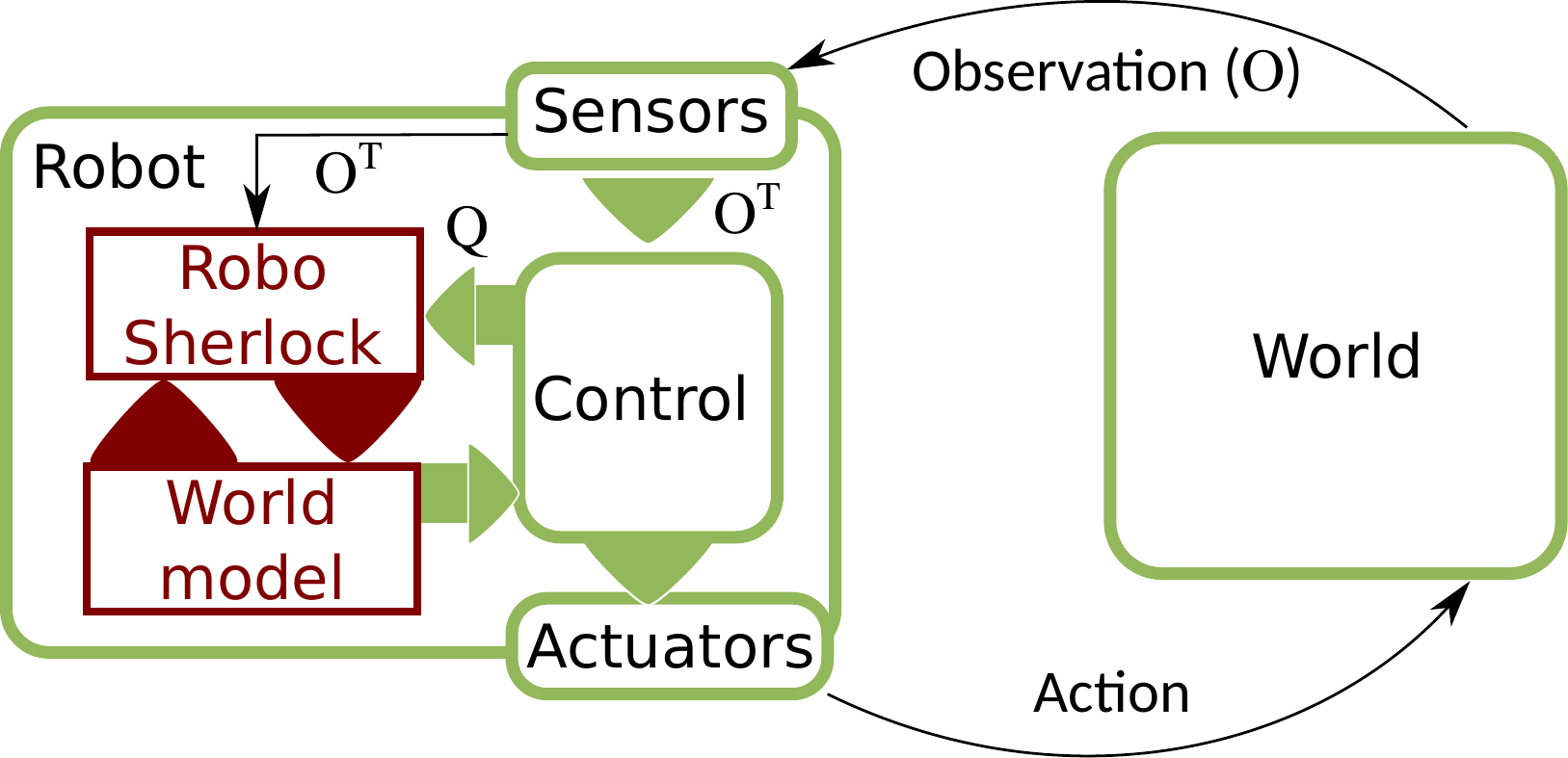}
    \caption{The classical perception action loop, with RoboSherlock's role in it highlighted}
    \label{fig:perc_action}
\end{figure}

\newtext{In this section, we phrase the perception problem for everyday 
manipulation tasks as a knowledge-enabled query answering problem. To 
this end, we consider a robot model that is inspired by the concept of 
rational agents originally introduced by \citet{russell10aima}. The 
main concepts are depicted in Figure~\ref{fig:perc_action}. A 
\emph{robot} is a physically embodied agent located in an 
\emph{environment} (which we also call a \emph{world}) and it perceives 
the world through its sensors and it is able to manipulate the world by 
its actuators. Let us denote the set of possible states of the world by 
$\mathcal{X}$. Typically, the state of the world is not directly 
accessible to the robot, but its sensors yield a filtered, noisy 
representation of a certain cutout of the world. The so-called 
\emph{perceptual filter} function determines how a world state is 
transformed into signals of the robots' sensors, $f_p: \mathcal{X} 
\mapsto \mathcal{O}$, where $\mathcal{O}$ denotes the set of possible 
sensor readings, which we also call \emph{observations}. Given an 
observation, the robot's control program is to decide on the next 
action to conduct, which is executed by the robot's actuators. Let 
$\mathcal{A}$ denote the set of possible actions, then the transition 
function $f_e:\mathcal{A}\times\mathcal{X}\mapsto\mathcal{X}$ specifies 
the effects that an action has on the environment, when being executed 
in a particular world state. The repeated execution of }
 
\begin{enumerate}

    \item \newtext{perceiving the state of the world}
    \item \newtext{selecting an action to conduct}
    \item \newtext{manipulating the world according to the action}

\end{enumerate}

\noindent\newtext{is called the perception-action loop. Let $T=\{1,\ldots\}$ 
denote a set of iterations (time steps) that the robot has executed the 
perception-action loop. Then, the robots' actions produce a sequence of 
world states $X^T:T\mapsto\mathcal{X}$, which are perceived by the 
robot as a sequence of observations $O^T:T\mapsto\mathcal{O}$, where 
$O^T(t)=f_p(X^T(t))$. The robot maintains an internal representation of 
the world, which is also called the robot's \emph{belief state}. Let 
$\mathcal{S}$ denote the space of all possible belief states. In every 
iteration of the perception-action loop, the belief state gets updated 
by the perception system, such that a trajectory 
$S^T:T\mapsto\mathcal{S}$ of belief states is generated over time. The 
belief state is used by the robot's control program to select the next 
action to conduct, and in turn serves as a basis for formulating a 
query $Q$ that is issued to the perception system in order to populate 
the world model with task-relevant knowledge about the environment. Let 
$\mathcal{Q}$ denote the set of possible queries, and $Q^T:T\mapsto Q$ 
the sequence of queries that the control program asks over time.}

\newtext{The \emph{perception problem for everyday robot manipulation} can now 
be formulated as a function
\begin{align}
    \rsshort:\ \mathcal{S}\times\mathcal{Q}\times\mathcal{O}^T\mapsto\mathcal{S},\label{eq:perception}
\end{align}}
\newtext{\noindent such that the following conditions hold:}

\begin{enumerate}
    \item \newtext{$S(0)=\emptyset$}
    \item \newtext{$\rsshort(S^T(t),Q^T(t),O^t)=S(t+1), \forall t\in T$,}

\end{enumerate}

\noindent\newtext{where $O^t$ denotes the sequence of past observations 
up to time $t$. This allows reasoning about world states taking into 
account sensor readings from multiple timesteps, which may lead to more 
accurate estimates than considering only one shot observations. 
Examples of such approaches are Bayesian filtering models, such as the 
Kalman filter, SLAM~\cite{thrun05probabilistic}, or Kinect Fusion~\cite{Izadi11KinectFusion}. } 

\section{\robosherlock~Overview} 
 \label{sec:uima_perc}

 \newtext{Having defined our notion of robot perception for everyday 
 manipulation, we will, in this section, introduce the basic ideas of 
 \rs, a knowledge-enabled framework that leverages implementations the 
 \rsshort\ function in Equation~(\ref{eq:perception}) that scale 
 towards realistic sets of objects, environments, and perceptual tasks, 
 \robosherlock\ considers perception as \emph{content analytics in 
 unstructured data}, first introduced by~\citet{blodow14phd}. Content 
 Analytics (CA) denotes the discipline of applying methods from the 
 field of statistical data analysis to large amounts of data in order 
 to extract semantically meaningful knowledge from those. The data are 
 considered \emph{unstructured} since the structure of the data does 
 not reflect its semantics as it for instance does in database schemata 
 or spreadsheets. The paradigm of \emph {unstructured information 
 management} (UIM) offers an implementational framework for realizing 
 high-performance CA systems. Perhaps the most prominent example of a 
 UIM system is \emph {Watson}~\cite{FerrucciEtAl10aimag}, a question 
 answering 	system that has won the US quiz show \emph{jeopardy!}, 
 competing against the champions of the show and demonstrating an 
 unprecedented breadth of knowledge. In UIM, pieces of unstructured 
 data, such as web pages, text documents, audio files or images are 
 processed by a collection of specialized information extraction 
 algorithms, and each algorithm contributes pieces of knowledge with 
 respect to its expertise. Thereby, outputs of different algorithms are 
 allowed to be complementary, overlapping or even contradictory. Hence 
 subsequently, the collected annotations are rated and consolidated to 
 come to a consistent final decision.}

Images are perfect examples of unstructured data. They depict hierarchical structures of object constellations, objects, object parts and relationships between these components but they are simply stored as an array of pixels. Thus, the UIM paradigm of hypothesize, annotate, test-rank can be, with minor modifications, applied to streaming sensor data. \robosherlock\ operates based on this principle. It creates object hypotheses for pieces of sensor data that it believes to represent objects or object groups. Subsequent perception algorithms analyze these hypotheses and annotate their results as semantic meta data to the hypotheses with respect to their expertise. Further algorithms then test and rank possible answers to the given perceptual task based on the combination of sensor and meta data. To this end, the framework supports the application of multiple algorithms to various objects with different properties and the combination of their results. This is achieved in multiple ways: $\circ$ by providing means for the communication among expert methods; $\circ$ by providing infrastructure that supports reasoning about how results of different experts should be combined; $\circ$ by enabling the system to reason about its perceptual capabilities and decide which methods are the best for a certain perceptual sub-task.

\robosherlock\ is used to perceive the information that a robot needs in order to accomplish its manipulation tasks. Perception tasks can be formulated and issued to \robosherlock\ in the following ways:

\begin{enumerate}
    \item \textbf{detect \textit{obj-descr}} asks \robosherlock\ to detect objects in the sensor data that satisfy the description \textit{obj-descr} and return the detected matching \textit{object hypotheses}. 
 
	\item \textbf{inspect \textit{obj-hyp} \textit{attributes}} asks the 
	perception system to examine a given hypothesis \textit{obj-hyp} in 	
	order to extract additional information as requested by \textit{attributes} 
	and add the information to \textit{obj-hyp}. 

	\item \textbf{track/scan \textit{task-descr}} runs continuous perception 
	for a description defined in \textit{task-descr}; These are perception 
	tasks that typically require a stream of images, i.e. track an object while 
	performing a manipulation task.
\end{enumerate}

\robosherlock\ is designed to accomplish such perception tasks in complex scenes that include objects with different perceptual characteristics. To do so, objects are perceived perceives taking the scene context into account and different perception methods are employed and reasoned about in order to decide which methods to apply to which objects. \robosherlock\ can use background knowledge to simplify perception tasks, for example when the object it is looking for has a salient distinctive characteristic, or in order to interpret the perception results by inferring whether or not the object could hold a liter of some liquid. In order to improve robustness \robosherlock\ can also reason about how to combine the results of different perception methods. 


\section{Conceptual overview}
\label{sec:perc_as_uim}


\robosherlock\ has been designed with two major implementational aspects in 
mind: (1) it does not replace any existing perception system or algorithm but 
rather enables easy integration of previous work in a unifying framework that 
allows these systems to synergistically work together and (2) new methods can 
be easily wrapped into \robosherlock\ processing modules to extend and improve 
existing functionality and performance.

\newtext{Conceptually there are two main parts of the framework. One for query interpretation and planning perceptual processes and one for processing the raw data, a collection processing component. The former is our contribution to the problem of content analytics while the later is based on key concepts of UIM that we have adapted to fit the needs of robot perception. These two parts are depicted in Figure ~\ref{fig:rs_conceptual_view}.}

The key concepts that \robosherlock\ builds on are:
\begin{itemize}
    \item the \textit{CAS} with the views to be analyzed, which is the common 
	data structure and acts as a short term memory of the \robosherlock\ 
	system; 
    \item a \textit{type system} that acts as a common language, allowing 
	communication between components;
    \item the \emph{analysis engines (AEs)} the core processing components of the framework implementing parception algorithms and complete perception pipelines. They share and operate on the CAS by generating, interpreting, and refining hypotheses;
    \item \newtext{\textit{Collection readers} for interfacing with the sensor models of the system and initializing the CAS;}
    \item \newtext{\textit{CAS consumers} for solving inconsistencies merging results and updating the world model;}
    \item a \newtext{\textit{belief state representation} that represents our current understanding of what the world looks like and allows for processing mechanisms that the AEs can use as resources for reasoning about the objects 
    and scenes they interpret.}
\end{itemize} 

We will present in detail each of these concepts. An example implementation of the pipeline planning that is used in our systems is given in Chapter~\ref{sec:pipeline-gen}. Conceptually planning a perceptual process involves generating all possible processing pipelines and choosing the ones that can satisfy our query. If $Q=\emptyset$ a predefined sequence of perception algorithms can be used.


\subsection{\newtext{Belief state representation\\(formerly: The type system)}}

\label{ssec:type-system}

The robot's belief state is a formal representation of the state that 
the robot believes the world to be in. It is therefore crucial that the 
representational formalism is expressive enough to capture the essence 
of the world at a sufficiently high level of detail. At the same time, 
one should be able to reason about aspects of the model in order to 
derive new knowledge about the world that is only implicitly entailed.

In \rs, we advocate the use of \emph{description logics} (DL) as the 
representation language for two reasons: (1)~DL provides an intuitive, 
standardized and clean way of constructing models of the real world and 
(2)~DL has well-established implementations used in both the knowledge 
representation and the AI-based robotics and communities. In our 
implementation, we use the KnowRob~\cite{tenorth2011phd} knowledge base 
as a basis.

\emph{Description logics} is a subset of first-order logic that 
facilitates ontological engineering of world representations. An 
ontology in DL consists of two main components: a \emph{TBox} and an 
\emph{ABox}. The TBox defines the terminological building blocks a 
belief state may be composed of. It contains a set of \emph{type 
symbols} $\top$ and universal rules how they relate to each other. 
Examples of type symbols are \textit{Container}, \textit{Cup}, 
\textit{Milk}, or \textit{Number}. One of the most prominent relations 
is the \emph{subsumption}, wich hierarchically arranges the types in a 
taxonomy, i.e. $\sqsubseteq\ \subseteq\top\times\top$. Propositions like
\begin{align*}
\textit{Cup} & \sqsubseteq\textit{Container}\\
\textit{Bowl} & \sqsubseteq\textit{Container}\\
\textit{Container} & \sqsubseteq\textit{PhysicalThing}\\
\textit{Milk} & \sqsubseteq\textit{Liquid},
\end{align*}

\noindent for instance, state that the types \textit{Cup} and 
\textit{Bowl} both are specializations of the type \textit{Container}, 
which in turn is a specialization of the type \textit{PhysicalThing}, 
and that \textit{Milk} is a specialization of the \textit{Liquid} type.
Additional relations can be used to introduce more detailed type
definitions, such as 
$$\textit{Cup} \doteq \emph{Container}\,\sqcap\exists\,\textit{holds.Liquid}\,\sqcap\exists\,\textit{has.Handle},$$

\noindent which defines the concept of a cup as a the  intersection of 
the concept of a container that has a handle and holds some liquid. We 
will refer to the TBox of the belief state also as the \rs\ \emph{type 
system} in this paper. A specific instantiation of a world modeled in 
DL is stored in the ABox, which contains symbols referring to 
individuals in the world. Concept assertions and relations can be used 
to assign certain individuals one or more types from the TBox, like 
$$\textit{Cup}(c)\text{ and }\textit{Milk}(t),$$

\noindent if an entity $c$ is an instance of the concept \textit{Cup} 
and a different entity $t$ is an instance of the concept \textit{Milk}. 
If the relation $\textit{holds}(\cdot,\cdot)$ holds for the pair 
$\langle c,t \rangle$, then the assertion $$\textit{holds}(c,t)$$ must 
hold. A comprehensive treatise of description logics is beyond the 
scope of this paper. For an excellent introduction, we refer the reader 
to \cite{Rudolph2011}.

It is important to note that all data structures and algorithms 
integrated in \rs\ are required to be modeled in the belief state. This 
(1) guarantees the interchangeability of data in a common language 
between components with unique, well-defined semantics, and (2) in 
turn, allows explicit reasoning about the perceptual capabilities of 
the robot, which is a precondition for the dynamic generation of 
processing pipelines that satisfy a query. For example,

\begin{align*}
\textit{ClassificationAnnotation}\doteq&\,\exists\,\textit{classLabel.}\top\\
                                 &\sqcap\exists\,\textit{classConfidence}.\mathbb{R}\\
                                 &\sqcap\exists\,\textit{classifierName.String}
\end{align*}

\noindent defines the result of a classification algorithm as an entity 
with attributes \textit{classLabel}, \textit{classConfidence}, 
\textit{classifierName}. The type system may further specify that the 
\textit{classConfidence} must be filled with exactly one floating point 
value and \textit{classLabel} itself is a type. For example, the type 
\textit{ClassificationAnnotation} is a sub-type of 
\textit{SemanticAnnotation} which in turn is a sub-type of 
\textit{FeatureStructure}. By storing this taxonomy in a centralized 
knowledge-base, we can easily ask for a list of AEs that have a 
specific type of annotation as output. For the perceptions algorithms, 
pre- and post-conditions in the TBox specify under which circumstances 
algorithms may be applied and which results they produce. In 
Section~\ref{sec:pipeline-gen}, we will describe the pipeline 
generation in greater detail.

The \robosherlock\ type system defines types for each type of 
annotation and object hypotheses. Besides the types that 
serve as blueprints for objects that hold interpretations of the raw 
data, we also define types for low-level vision data structures (point 
clouds, images, regions of interest, rigid transforms).


\subsection{A meta language for formulating\\ perception tasks}
\label{sec:meta_language}


\newtext{Having the belief state represented in a formal, logic-based language 
allows to answer queries about different aspects of the world at 
runtime. A \emph{query} is a statement in a formal declarative language 
that describes conditions that entities in the belief state need to 
satisfy in order to be among the query result. The result of a query 
thus is a set of tuples that match that description. For a robot, such 
reasoning capabilities are extremely important as it needs to be able 
to retrieve this kind information on demand, such as ``a free location 
on the table to put down an object'', ``an empty container object that 
is able to hold at least two liters of water'', or ``the turning knob 
on the oven''. For our belief state in DL, we can use tuple calculus 
known from relational algebra~\cite{codd1970relational}. A generic 
query in relational calculus has the form $\{t\,|\,\Psi(t)\}$, where 
$t$ denotes a variable representing a candidate tuple and $\Psi(t)$ is 
a statement that formulates the requirements that $t$ needs to satisfy 
in order to qualify as a result. A query for ``an empty container that 
is able to hold at least 2 liters of water'' can be formulated as 
follows:} 
\begin{align}
    \{c\,|\,c\in\textit{Container}&\land c.\textit{capacity}\ge2\nonumber\\
    &\land\lnot\exists s\in\textit{holds}(c.\textit{id}=s.\textit{id})\},\label{eq:detect-query}
\end{align}

\noindent and a query for ``the pose of the turning knob on the oven'' 
can be formulated as
\begin{align}
    \{[n.\textit{pose}]\,|\,n\in\textit{Knob}&\land \exists o\in \textit{Oven}\land o.id=\text{`oven-01'}\nonumber\\
    &\land o\in\textit{has}(o.\textit{id}=n.\textit{id})\}.\label{eq:inspect-query}
\end{align}

\newtext{We can interpret queries to the belief state of the robot also queries 
to the perception system. In query (\ref{eq:detect-query}), we query 
for all tuples of the type \textit{Container} that have a capacity of 
at least two liters and are not in a \textit{holds} relation with 
anything else. This naturally corresponds to a \textbf{detect} 
perception task. In query (\ref{eq:inspect-query}), we query for an 
object of type \textit{Knob}, which is in a \textit{has} relation with 
a specific object of type \textit{Oven}, and we return the 
\textit{pose} attribute of that object. Query~(\ref{eq:inspect-query})
thus corresponds to an \textbf{inspect} query.}

In the following, we will sketch a machine interpretable meta language 
in which we can state a subset of perception queries for robots 
performing manipulation tasks, whose semantics is defined in terms of 
queries in relational tuple calculus as described above. The perception 
task language enables us to better understand the capabilities of 
perception algorithms and systems. We can use the language to state the 
tasks that perception algorithms can accomplish. We propose to state 
perception tasks in a symbolic language that consists of \textit{terms} 
and \textit{perception tasks}. 

The terms of the language are \textit{object descriptions}, 
\textit{object hypotheses} and \textit{task descriptions}.  Using such 
descriptions we can describe a red spoon as \textit{(an object 
(category spoon) (color red))}. 

\strike{Object hypotheses are data structures that represent hypotheses 
about detected objects, object groups, parts, etc. that can be passed 
around in the system to exchange information between different 
algorithms.} Perception tasks are then formulated as one of two 
operations on object descriptions and hypotheses:

\begin{enumerate}
	\item \textbf{detect \textit{obj-descr}}, which asks the perception
	system to detect objects in the sensor data that satisfy the
	description \textit{obj-descr} and return the detected
	\textit{hypotheses}.
	\item \textbf{inspect \textit{obj-hyp} \textit{attributes}}, which
	asks the perception system to examine a given hypothesis
	\textit{obj-hyp} in order to extract additional information as requested by \textit{attributes} and add the information to
	\textit{obj-hyp}.
	
\end{enumerate}

\noindent In more detail, and object detection task has the form:
\begin{tabbing}
	(\textit{detect} ($\langle$\textit{det}$\rangle$ \= 
	$\langle$\textit{type}$\rangle$
	\\
	\> ($\langle$\textit{attr$_{1}$}$\rangle$
	$\langle$\textit{val$_{1}$}$\rangle$) \\
	\> \ldots \\
	\> ($\langle$\textit{attr$_{n}$}$\rangle$
	$\langle$\textit{val$_{n}$}$\rangle$)))
\end{tabbing}

\noindent where $\langle$\textit{det}$\rangle$ is the determiner of the description, which can be one of the key words \textit{a(n)} or \textit{the}. If the key word is \textit{a(n)} than any hypothesis is accepted. If the determiner is \textit{the} there is assumed that exactly one hypothesis is generated. \textit{the} is needed to express a perception task such as find \textit{my} cup. If there are more than one objects matching the description of my cup this ambiguity should be detected \newtext{by the high level control system and handled}. $\langle$\textit{type}$\rangle$ can be
\textit{object}, \textit{object-part}, and
\textit{scene}. \textit{object} asks for objects and object groups that cannot be further segmented by the perception system. For example, segmentation algorithms typically cannot segment a set into
the cup  and the spoon inside it. \textit{object-part} specifies parts such as the handle of a spatula, and
\textit{scene} for constellations of objects. The determiner of the description is followed by optional attribute value pairs. Some attributes can be abstract characteristics such as affordances (\textit{graspable}, \textit{can-hold}). In addition, we can state the locations where the objects can be found such as in a container, on a surface, or relative to a reference object such as in front of the cup.

The second category of perception tasks are examination tasks that allow the robot to perceive additional information about detected objects. Formally, an examine query takes the form:

\begin{tabbing}
	(\textit{inspect} \=$\langle$ \textit{\#uid} $\rangle$ \\
	\> $\langle$\textit{:attr$_{1}$}$\rangle$ 
	$\langle$\textit{:attr$_{2}$}$\rangle$ \ldots 
	$\langle$\textit{:attr$_{n}$}$\rangle$)
\end{tabbing}

\noindent where \textit{\#uid} is a unique identifier of an object hypothesis 
and $\langle$\textit{:attr$_{1}$}$\rangle$ 
$\langle$\textit{:attr$_{2}$}$\rangle$ \ldots 
$\langle$\textit{:attr$_{n}$}$\rangle$ is the list of attributes we want to  
examine for. Object attributes that can be inferred from perception algorithms 
are for example \textit{pose}, \textit{2D/3D model}, \textit{object state}, or 
\textit{grasp points}. 

The detection task serves as the base query of the language. 
A special category of perception tasks are ones that we call compound queries. These queries use the detection tasks as a sub-query in order to achieve a more complex behavior. Such tasks can be ones that ask the perception system to start a continuous perception task. The description can contain information such as a command for starting and stopping a task, description of an object, or a given hypothesis.

\begin{tabbing}	
	(\textit{track/scan/...} \=$\langle$\textit{det}$\rangle$ $\langle$\textit{type}$\rangle$ (\\
	\> detect (\=$\langle$\textit{det}$\rangle$  
	$\langle$\textit{type}$\rangle$
	\\
	\>\> ($\langle$\textit{attr$_{1}$}$\rangle$
	$\langle$\textit{val$_{1}$}$\rangle$) \\
	\>\> \ldots \\
	\>\> ($\langle$\textit{attr$_{n}$}$\rangle$
	$\langle$\textit{val$_{n}$}$\rangle$)))
\end{tabbing}

\noindent These tasks include for example the tracking of an object while performing a manipulation task, or the scanning of a region for fusing results from multiple images.

\begin{table}[ht]
	\begin{center}
		\begin{tabular}{|l|p{6cm}|}
			\hline
			\textbf{Attribute} & \textbf{Description} \\
			\hline\hline
			\textit{shape} &  semantic shape label\\
			\hline
			\textit{color} & semantic color label\\
			\hline
			\textit{type} & type of an object e.g. super-class in taxonomy, or object affordance\\
			\hline
			\textit{location} & semantic location of an object (e.g. on table top, in a drawer)\\
			\hline
			\textit{class} & result of a classification be it classification of object instances or class\\
			\hline
			\textit{pose} & specify a pose\\
			\hline
			\textit{cad-model} & specify a CAD model to fit\\
			\hline
			\textit{obj-part} & specify the part of an object to detect (e.g. handle, opening etc.)\\
			\hline
			... & ... \\
			\hline
		\end{tabular}
	\end{center}
	\caption{List of attributes currently implemented in \robosherlock}
	\label{tab:attributes}
\end{table}

The list of attributes is not meant to be exhaustive and Table~\ref{tab:attributes} contains only those classes that are most important for object detection and perception in autonomous robot manipulation. Depending on the environment in which a robot is deployed, the metalanguage can be extended to handle  attributes that are specific to the tasks that need solving.

Using the introduced metalanguage we can state the task to find a flat object that has the color black and is located in a specific drawer the 
following way:

\begin{small}
	\begin{tabbing}
		(detect (an \= object \\
		\> (shape flat) (color black) \\
		\> (loca\=tion in (a container (category drawer\#3)))))
	\end{tabbing}
\end{small}

\noindent and examine it for its pose and grasp points with:

\begin{small}
	\begin{tabbing}
		(inspect \#obj\_id :pose :grasp-points)
	\end{tabbing}
\end{small}

Or, we can state the perception task to track an object of a certain type with
\begin{small}
	\begin{tabbing}
		(track (an \= object (type `Spatula`) \\
		\> command `start`)) 
	\end{tabbing}
\end{small}

In order to generate context-specific perception plans that invoke specialized  perception algorithms (experts) as plan steps and to reason about the results  obtained, terms of the metalanguage are interpreted and reasoned upon by the perception framework. The query itself is represented in the type system and stored in the CAS during execution.

\begin{figure}[t]
	\centering
	\includegraphics[width = 0.99\columnwidth]{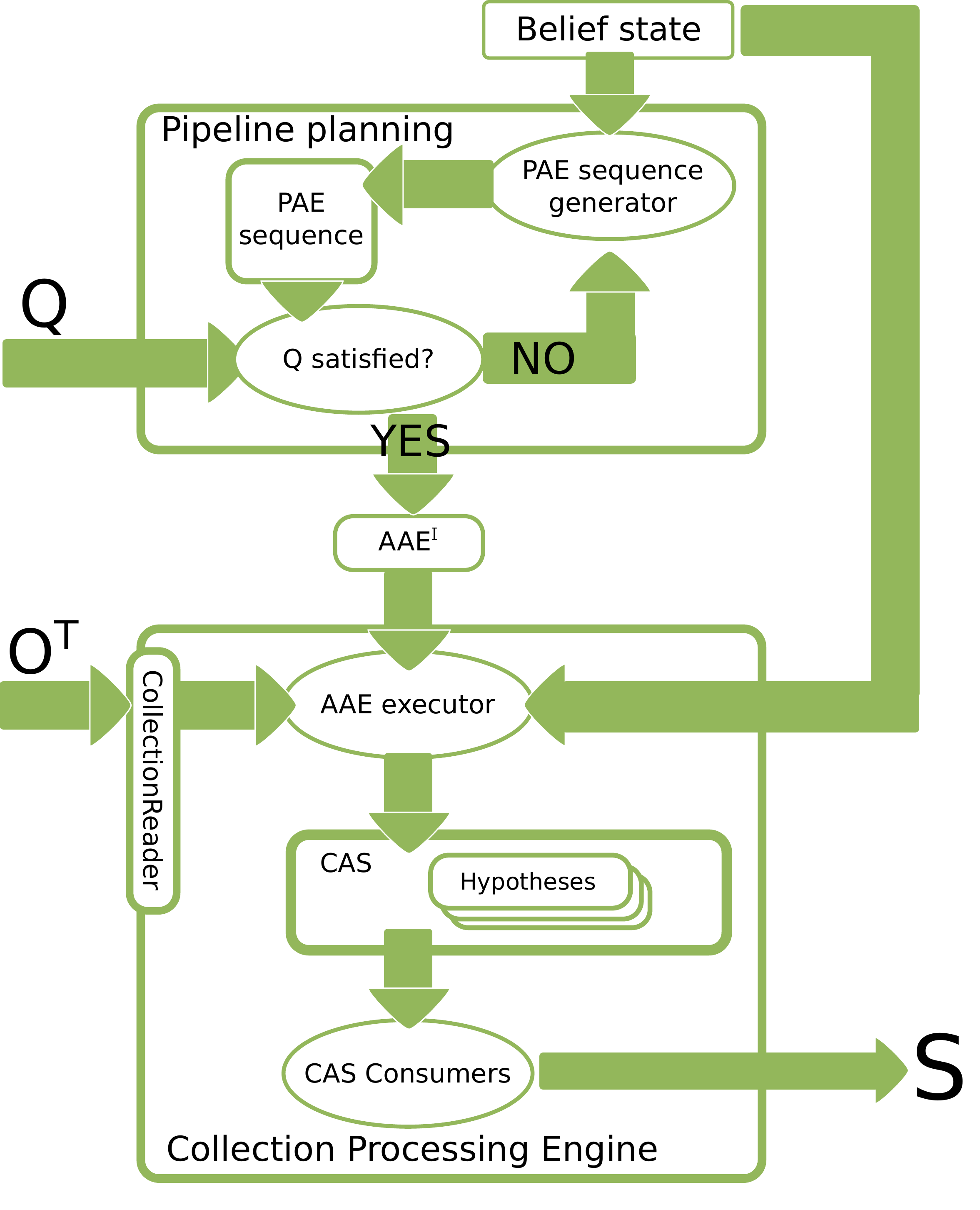}
	\caption{High level conceptual overview of an execution cycle in RoboSherlock}
	\label{fig:rs_conceptual_view}
\end{figure}

\subsection{\newtext{The Collection Processing Engine}}
\label{sec:analysis-engines}	

Figure~\ref{fig:rs_conceptual_view} presents a conceptual overview of how a single processing cycle in \rs\ looks like. Before describing such an cycle we first introduce the concepts necessary to understand the behavior.

We start by defining the CAS, the central data structure that is used by the components of the \robosherlock-based perception systems in order to communicate with each other.
A CAS consists of:  
\begin{itemize}
	\item \myem{\sofa s} (Subjects of Analysis) the raw sensor data that is to 
	be interpreted (e.g. RGB or depth image), which in our model is equivalent to the observations $O^T$ 
	\item and the \textbf{hypotheses} that are generated during a processing cycle.
\end{itemize}

A hypothesis represents a certain region in our observations and a set of annotation attached to this region. A region in an observation is a set of indices of the vector of observations, e.g. pixels or 3D points. An annotation is an ABox representing a fragment of the belief state.

Let $\mathcal{P}(\mathbb{N}^+)$ be the power set of pixels of an image or points of a cloud, and then the a hypotheses $\mathcal{H}yp$ is defined as 
\[ \mathcal{H}yp := \mathcal{P}(\mathbb{N}^+) \times \text{ ABox}.\]

\noindent An exemplary hypothesis of a box-like red object which has a pose 
and a 3D feature descriptor using the description logic introduced earlier would have the form:

\begin{equation*}
\begin{split}   
 Thing(h_1) {}& \land shape(h_{1},box) \land color(h_{1},red) \\
 & \land pose(h_{1}, [x,y,z,r,p,y]) \land vfh(h_{1},[...])
 \end{split}
\end{equation*}

\begin{figure*}[h]
	\centering
	\includegraphics[width=0.99\textwidth]{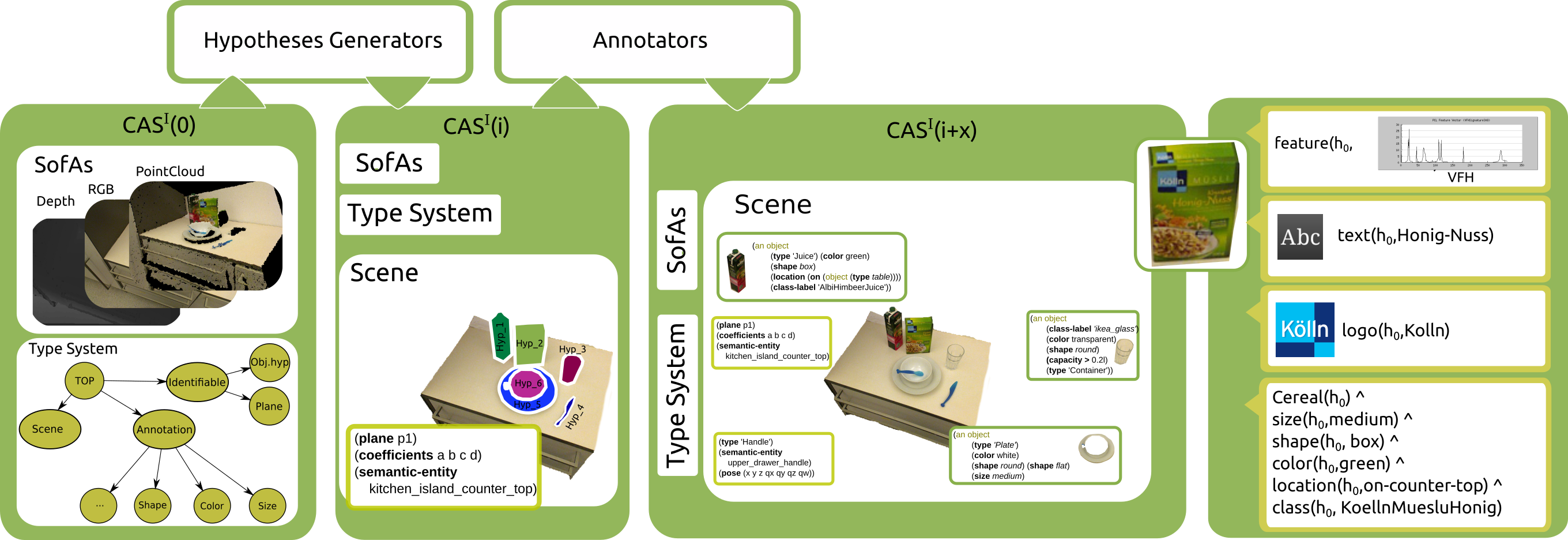}
	\caption{The common analysis structure of a kitchen breakfast scene as it gets filled by annotators during one execution cycle.	}
	\label{fig:cas}
\end{figure*}

\begin{mydef}
	CAS\\ 
	Now we can define what a CAS is. Let $\mathcal{P}(\mathcal{H})$ the power set of all hypotheses. The space of all CAS-es is defined as: 
	\[CAS := \mathcal{O} \times \mathcal{P}(\mathcal{H}yp)\] 
	The CAS is re-initialized with every new observation $O^T(t)$ and enriched by components in the framework. The result sequence of the CAS is thus defined as: 
	\begin{align}
	  CAS^I : I \longmapsto CAS 	
	  \label{eq:CASI}
	\end{align}
	where $I$ is a set of indices that corresponds to primitive analysis engines in the pipeline. 
\end{mydef}

\noindent Figure~\ref{fig:cas} depicts a CAS as it is produced and updated by \robosherlock\ analyzing a table scene. Right side of figure shows an object hypothesis that depicts a cereal-box. 

Now that we have introduced the data components of our framework, let us look at the processes that use these. Collection readers ($CR$s) are special purpose modules that are designed to read in the observation from the sensors and initialize the CAS. A collection reader can be defined as:

\[ CR: \mathcal{O^T} \longmapsto  CAS \hspace{1ex} s.t. \hspace{1ex} CR(O^t) = \langle O^t(t), \emptyset \rangle , \forall t \in T \] 

\noindent creating a CAS with the most recent observation and an empty set of hypotheses.

The core processing elements of \robosherlock\ are analysis engines. They share the CAS and operate generate, annotate and refining object hypotheses. Analysis engines can be divided into two categories: primitive AEs ($PAE$) and aggregate AEs ($AAE$). 

Primitive AEs can generally be split based on their capabilities into two categories: the first kind, called \emph{hypotheses generators} analyzes the observations and generates region of interests. The second one, called \emph{annotators}  annotates the generated regions of interest. Some primitive AEs do not fit into the one or the other category but combine both categories. An example for such a primitive AE is a CAD model fitter which generates a hypothesis and, if successful, annotates the hypothesis with a class label.
Conceptually both hypotheses generators and annotators are the same, both taking the CAS as an input and enriching it: 

\[ PAE: CAS \longmapsto CAS \]
 
Aggregat AEs solve more complex tasks. An AAE consists of an ordered list of primitive AEs. The primitive AEs can be run sequentially, in parallel, or flexibly, e.g. on-demand or event-driven. The planning of a perception pipeline for a query results in an ordered sequence of primitive analysis engines, $AAE^I$.  

\[ AAE^{I}: I \longmapsto \mathcal{PAE} \] 

\noindent where $I$ is the same set of indices as in~\ref{eq:CASI} and $\mathcal{PAE}$ is the set of all primitive analysis engines.

An AAE execution component takes the initialized CAS with the observations and runs the sequence of $AAE^{I}$ on it, outputting the final version of the CAS, that contains a partial semantic description of the world at time $t$. To relate this back to enrichment of the CAS during execution consider the following:

\[ CAS^I(i+1) = AAE^I(i)(CAS^I(i))\]

\noindent meaning that in order to obtain the next CAS in our sequence, we have to execute the current PAE on the current CAS.

In order to update the world model these descriptions need to be incorporated in to the world model $S$. Since annotators might employ heuristic interpretation methods or are more or less reliable and accurate, the set of annotations is allowed to be inconsistent or contradictory. Inconsistencies and erroneous annotations are handled by subsequent reasoning and hypothesis testing and ranking using components called \textit{CAS Consumers} (CC):
	\[ CC: CAS \longmapsto S \]

With these definition we can now define an iteration of a collection processing engine in \robosherlock\ as 
\[ (CC\circ S \circ AAE(i) \circ \dots \circ AAE(1) \circ  CR )(O^t, Q) = S^T(t)\]

\noindent that is the composition of a collection reader, the seqence of PAEs, the world model and the CAS consumer for a sequence of observations and a query $Q$.


%

\section{Implementation of a perception system}
\label{sec:implementation}

\begin{figure*}[t!]
	\begin{center}
		\includegraphics[width=\textwidth]%
		{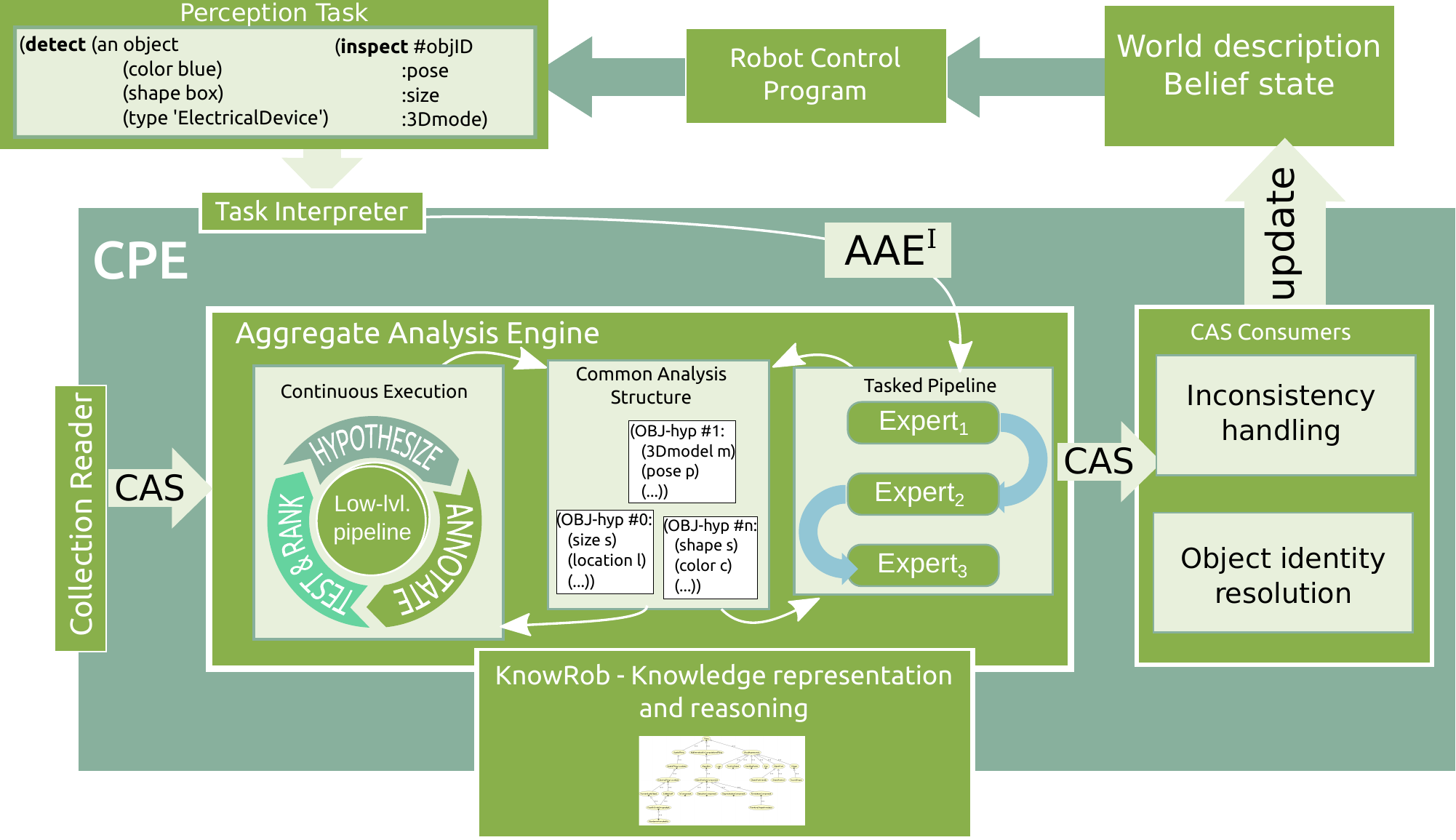}
	\end{center}
	\caption{High level overview of the~\robosherlock\ framework: the system 
		components and the most important interactions}
	\label{fig:system_overview}
\end{figure*}

\newtext{Based on the conceptual description of the framework, we now present an example perception system that is used in the context of everyday manipulation tasks performed in human environments. We give examples of PAEs implemented and how we can further categorize them based on their generality. The AAE that is formed is particularly interesting because of it's continuous nature that does not require a query to run, enabling a preparatory perception pipeline to pervasively gather information about the environment. We describe two approaches for CAS consumers, one for aggregating results over time and one for solving inconsistencies in hypotheses annotations. We conclude the Section by detailing how our world model ($S$) is implemented, how it is used to plan perception pipelines and exemplify reasoning tasks that are implemented.}

Figure~\ref{fig:system_overview} shows the high-level organization of this system. 
\robosherlock\ is implemented using the Apache UIMACPP framework\footnote{https://uima.apache.org/doc-uimacpp-huh.html} and the ROS middleware, interfacing with sensors and other software components happens through typical ROS interfaces. An observation $O^T(t)$ in our case corresponds to a set of images taken by the robot at timestamp $t$. The CollectionReader reads the images together with their meta data and initializes the CAS.

\newtext{The internal data processing cycle has two distinct components: a continuous execution component and a tasked execution component. Both follow the \textit{hypothesize, annotate, test and rank} paradigm of UIM. The continuous executions pervasively detects object hypotheses and annotates the hypotheses with numeric (key-points, descriptors) and semantic (shape, color, size) data in a fast-loop~(10Hz). Formally, this would be described through Equation(~\ref{eq:perception}) such that  $\mathcal{Q} = \emptyset$. The sequence of PAEs is pre-defined at start-up.}

Perception tasks are passed to the system in the form of perception queries that are handled by the task interpreter. As we already mentioned, generating all possible sequences of PAEs is not feasible. Instead we interpret the query and decide whether the assertions made by continuous component are sufficient to answer it. If this is not possible, the system plans a new pipeline specific to the query using the implemented reasoning mechanisms and executes that.

After the initial CAS containing the raw data is passed to the AAE, the sequence of PAEs is executed.  PAEs are mainly perception algorithms performing various tasks. In the scenarios we are considering here this sequence starts with the execution of object hypotheses generators. These PAEs analyze the images and  detect regions that correspond to objects and object groups in the environment. Subsequent perception algorithms (specialized routines called \emph{annotators}) analyze these hypotheses and annotate them with symbolic or numeric information. The results are asserted into the CAS as 
annotations of hypotheses. In Figure~\ref{fig:example_scene} the object hypotheses generators have detected raw point clusters that are indicated by the overlaid contours.

\newtext{Although all PAEs conceptually are the same, in practice we can distinguish between them based on what parts of the CAS are they enriching. We have already differentiated between hypotheses generators and hypotheses annotators. We can also categorize them based on their applicability: general-purpose ones that can be run for any object hypothesis and task-specific ones that need some specific world stat in order to be applicable. This distinction is easiest illustrated considering the scene depicted in Figure~\ref{fig:example_scene}. A general purpose PAE would be a visual key-point extractor, or a color histogram estimator for the objects in the scene. These algorithms can be run nu matter what the  hypotheses and return annotations that can be useful for further PAEs down the execution cycle. A perception task-specific annotator would be one that can only run and return positive results when task conditions are met. For example a drawer handle detection algorithm in the considered scene that identifies handles as 3D lines in the depth image. This annotator can only work and can only be used if the drawer handles in our world instance ($ABox$) meet the requirements. As another example, typically, the volume of an  objects is not of interest unless a query specifically ask for it. Testing such a property is done through the application of knowledge processing and reasoning techniques, detailed in Section~\ref{sec:knowledge_integration}, and is only performed if a task demands it.}

\newtext{After every processing cycle the CAS is transmitted to the consumers where further algorithms can analyze the generated hypotheses, filter them, merge them, solve inconsistencies and finally assert them into the current belief about the world. Having such methods is optional but recommended. If an application does not demand a consistent world model, the hypotheses can be directly integrated into the world model and sent back as final results. In subsection ~\ref{sec:merging} we briefly present two of our previously published approaches for ensuring consistency.} 

\begin{table*}[htb]
	\centering
	\footnotesize{
		\tymin=3cm
		\begin{tabulary}{\textwidth}{|L|L|L|}
			\hline
			\textbf{AE} & \textbf{Type}  & \textbf{Description} \\
			\hline
			Color & Annotator & Returns semantic color annotation based on 
			color distribution in HSV color space. Also annotates the 
			hypotheses with a color histogram\\
			\hline
			PointCloudClusterExtractor & Hypotheses generator & Segmentation of 
			table top scenes based on euclidean distance\\
			\hline
			ImageSegmentation &  Hypotheses generator  & Binary image 
			segmentation\\
			\hline
			PCLDescriptorExtractor & Annotator  & Extracts 3D features 
			implemented in PCL (VFH, SHOT etc.)\\
			\hline
			KeyPointExtractor & Annotator & Extracts keypoints and calculates 
			keypoint-descriptors using OpenCV (FREAK, FAST, SIFT, SURF, etc.)\\
			\hline
			TransparentSegmentation & Hypotheses generator &  Segmentation of 
			transparent objects as described by ~\cite{Lysenkov-RSS-12}\\
			\hline
			RSSvm, RSRf, RSKnn, $\ldots$ & Annotator & Support vector machine , Random Forrest, k-NN classifier wrappers etc. 
			classification annotators from OpenCV\\
			\hline
			CaffeAnnotator & Annotator & Extracts deep learning features using 
			pre-trained models in  Caffe~\cite{jia2014caffe}\\
			\hline
			3DGeometry & Annotator & Estimates a pose based on a 3D oriented 
			bounding box. Also classifies objects into \textit{small} or  
			\textit{big} depending on 3D volume\\
			\hline
			Goggles & Annotator & Sends the image of a region of 
			interest of a hypotheses to the Google Goggles servers and parses 
			the answer to extract text, logo, and texture 
			information~\cite{blodow14phd} \\  
			\hline
			PrimitiveShape & Annotator  & Fits lines and circles to 3D point 
			clusters projected onto the 2D plane  using 
			RANSAC \cite{goron12robotik}. Values returned: \textit{box, round, 
			flat}\\ 
			\hline
			LineMod & Annotator \& Hypotheses generator & Matches each object 
			hypothesis to a set of object models that the robot should actively 
			look for using the Linemod 
			algorithm~\cite{hinterstoisser2011linemod}. \\  
			\hline
			SimTrack & Annotator \& Hypotheses generator \& Tracking. & Fits a 
			CAD-model from a query to the current scene and tracks it (only 
			works for textured objects)~\cite{pauwels_simtrack_2015}. \\  
			\hline
			TemplateAlignment & Annotator & Fits a CAD-model from a query to 
			a hypothesis that was classified as the CAD 
			label~\cite{tombari10clutter} \\  
			\hline
			SACmodel & Annotator & Fits parametric models to 
			objects if the query demands\\  
			\hline
			RegionFilter & Hypotheses generator & Interprets object positions 
			in terms of a semantic environment 
			map~\cite{iros12semantic_mapping}. Filters \emph{depth}- and 
			\emph{RGB}-image leaving only pixels in a region of interest. \\  
			\hline
		\end{tabulary}
	}
	
	\caption{Some of the analysis engines implemented in \robosherlock}
	\label{tab:aes}
\end{table*}

\subsection{Hypothesis Generation}

Perception algorithms and existing perception systems are wrapped as PAEs. These assert facts into the knowledge base and reason about their own capabilities as well as the task that is currently being executed. Table~\ref{tab:aes} presents some of the commonly used AEs that are implemented and offers a short description of what they do. The majority of these AEs wrap 
around perception algorithms from PCL~\cite{Rusu_ICRA2011_PCL}, OpenCV~\cite{opencv_library} or ROS packages that are commonly used in robotics, most of which can be assigned to one of the two categories of AEs: hypotheses generators or hypotheses annotators. 

In most cases hypotheses generators are specialized segmentation algorithms that find possible locations of items, and deal with objects that exhibit different perceptual characteristics such as ordinary objects of daily use, flat objects, pieces of paper or small shiny objects such as knives and forks. 

These methods are complemented with mechanisms to combine their results in a 
consistent manner including the following ones:

\begin{itemize}
	\item attention mechanisms that detect points of interest in pixel   
	coordinates in order to create regions of interest (points and extents) in 
	the camera frame.
	\item image segmentation algorithms (e.g. color-based) can generate masks 
	or region maps, referencing the respective part of the image.
	\item point cloud segmentation relying on supporting planar structures can 
	generate index vectors.
\end{itemize}

Note, that most of these types can be converted between each other e.g. 
projecting a point cluster from a point cloud into a camera image, or 
transforming an image region to a grasping pose in robot-local coordinates. 
This allows the retrieval of the camera image region of interest corresponding 
to a 3D point cluster, enabling the combination of image analysis annotations 
and point cloud processing. The PAEs most commonly used in \robosherlock\ for 
hypotheses generation are Euclidean clustering in 3D space to find 3D clusters and a color based segmentation in order to find objects that are missed by the 3D clustering (e.g. flat cutlery in a table setting). Having better segmentation algorithms (e.g. ones that deal with cluttered scenes) can furthermore improve the capabilities of the system. 

\subsection{Object Annotators}

Object annotators are the subclass of PAEs that enrich the hypotheses through the generation of annotations. In general, annotators wrap existing perception algorithms and result in numerical or symbolic values. There is a large variety of PAEs that can be implemented, we present a subset of these that we found important for everyday manipulation tasks.

For instance, the location annotator uses a 3D semantic map of an environment~\cite{iros12semantic_mapping} to annotate an object cluster with a semantically meaningful object location (e.g. ``on top of counter\_top''). 

The PCLDescriptorExtractor can process any point cluster (with estimated normals) and compute any open source 3D feature implemented in the point cloud library PCL (VFH, Spin images, RIFT, SHOT etc). Color, size or primitive shape annotators compute symbolic values for the hypotheses they are processing. All of the aforementioned AEs, can be considered as high-frequency AEs, and run in the continuous component, purely because of their execution times (order of milliseconds). 

On the other hand there are annotators like the one wrapping Google Goggles, a 
web service and smart phone app allowing the analysis of an image. Google 
Goggles generates a highly structured list of matches including product 
descriptions, bar codes, logo/brand recognition, OCR text recognition or a list 
of similar images. Some annotators wrap around existing perception frameworks, like 
BLORT~\cite{morwald2010}, Moped~\cite{Collet2011} Line-mod or SimTrack~\cite{pauwels_simtrack_2015}. These are object recognition or categorization engines, and are worth executing, when the task involves an object that has the respective model.

\subsection{\newtext{Merging hypotheses}}
\label{sec:merging}

\newtext{CAS consumers are important elements of the system. Their main purpose is to ensure the consistency of results. We briefly present two of our previous approaches, one for probabilistic merging of annotations that addresses the problem of often conflicting results of PAEs and one for object identity resolution during pick and place tasks for ensuring consistency over time of objects that a robot manipulates. It is out of the scopes of this article to offer detailed information and evaluation of these sub-sytems. For such detail we kindly refer the reader to our related publications.}

\subsubsection{Propagating uncertainty}

A particularly powerful method for resolving inconsistencies in \robosherlock\ is the application of first-order probabilistic reasoning described in previous work~\cite{icra14ensmln}.  An interesting issue is how can the system handle uncertainty, as specially in the case where we have multiple experts of the same nature (e.g. model fitting or shape classifiers). In ~\cite{icra14ensmln} we have shown how, often contradictory, results from different PAEs(referred to as experts) can be merged to form a consistent hypotheses through the use of probabilistic first order logic, based on past experiences. Using the so learned model one can leverage the power of the probabilistic framework to model which expert performs best for different situations and use the learned weights as priors for choosing the experts that are most probable to detect the queried items. Figure~\ref{fig:feat_queries} shows most probable features for some of the objects, where for instance we can see that Line-Mod has a high chance of detecting a pot, but for all other object it will return false positives.

\begin{figure}
	\centering
	\footnotesize
	\begin{tabular}{l|c|c|c|c}
		\hline
		\textbf{Ground Atom}             & \textbf{Cereal}        & \textbf{Chips}         & \textbf{Cup}           & 
		\textbf{Pot}           \\\hline\hline
		color(c,yellow)	&	\textbf{0.4264}	        & \textbf{0.3484}	&	\textbf{0.4422}	&	0.2936	
		\\\hline
		text(c,VITALIS\_A)	&	\textbf{0.623}	    & 0.0000	&	0.0000	&	0.0004	\\\hline
		logo(c,Kellogg's)	&	\textbf{0.3734}	    & 0.0000	&	0.0000	&	0.0008	\\
		linemod(c,Popcorn)	&	\textbf{0.7392}	    & 0.0006	&	0.0000	&	0.0010	\\\hline
		linemod(c,Pot)	&	0.0008	        & 0.0004	&	0.0004	&	\textbf{0.9994}	\\\hline
		linemod(c,PringlesSalt)	&	0.0002	& \textbf{0.4986}	&	0.0010	&	0.0006	\\
		shape(c,box)	&	\textbf{0.4806}	        & 0.3870	&	0.2810	&	0.3556	\\\hline
		shape(c,cylinder)	&	0.3722	    & \textbf{0.4540}	&	0.4010	&	0.4266	\\\hline
		shape(c,round)	&	0.3176	        & \textbf{0.4092}	&	\textbf{0.5182}	&	\textbf{0.4068}	
		\\\hline
		size(c,big)	&	\textbf{0.368}	            & \textbf{0.3442}	&	\textbf{0.3768}	&	\textbf{0.3292}	
		\\\hline
	\end{tabular}
	\caption{(Partial) probabilities for different queries about visual features conditioned on the object class. Excerpt 
		from previous work~\cite{icra14ensmln}}
	\label{fig:feat_queries}
	\vspace{-4ex}
\end{figure}

\subsubsection{Object Identity Resolution}

\newtext{Another highly relevant subsystem that is used as a CAS consumer is one for merging hypotheses over time and addressing the problem of object identity resolution during pick and place task in query driven systems~\cite{balintbe19amortized}. The subsystem keeps track of object that a robotic agent encounters during the execution of a task and solves new hypotheses to existing objects based on various distance metrics that are defined per annotation type. A peculiarity of the subsystem, is the use of background and task knowledge to reducde the number of false matches.}

\newtext{For example, given a semantic map of the environment~\cite{iros12semantic_mapping} and a localized robot, it is trivial to filter out any parts of the camera images that are not in scope of the current task, e.g. in a pick and place task only the source and destination regions are of interest. This reduces false detection and computational effort. Besides this in many robotic tasks the scenes are mostly static and successive frames are often similar and lead to no information gain, therefore skipping them offers more processing time for other tasks. Another common source for erroneous detections are motion blurred images. This happens if the camera or something in view is moving fast (like the manipulators of the robot). For the camera movement the pose of the camera in world space is tracked and movements bigger than a certain threshold are not considered. For the motion blur variations of the Laplacian detailed by Pech-Pacheco et. al~\cite{blur@icpr2000} can be used. Table~\ref{tab:episode_details} highlight the effects of these filters on four episodes of a pick and place task.} 

\begin{table}[t]
	\begin{center}
		\footnotesize
		\begin{tabular}{|l|l|l|l|l|l|l|l|l|}
			\hline
			& \multicolumn{2}{c|}{Ep.1} & \multicolumn{2}{c|}{Ep.2}& \multicolumn{2}{c|}{Ep.3}& \multicolumn{2}{c|}{Ep.4}\\
			\hline
			\# of Objs.  &\multicolumn{2}{c|}{9} & \multicolumn{2}{c|}{15}& \multicolumn{2}{c|}{20}& \multicolumn{2}{c|}{25}\\
			\hline
			duration &\multicolumn{2}{c|}{277(s)} & \multicolumn{2}{c|}{328(s)}& 
			\multicolumn{2}{c|}{510(s)}& \multicolumn{2}{c|}{520(s)}\\
			\hline
			\# of $Hyp$.  &\multicolumn{2}{c|}{130} & \multicolumn{2}{c|}{353}& \multicolumn{2}{c|}{694}& \multicolumn{2}{c|}{759}\\
			\hline
			\# pnp tasks &\multicolumn{2}{c|}{3} & \multicolumn{2}{c|}{4}& 
			\multicolumn{2}{c|}{6}& \multicolumn{2}{c|}{10}\\
			\hline
			filters & Off & On& Off & On& Off & On& Off & On\\
			\# objs. in bs.  & 15 & 9& 26 & 20& 49 & 28& 54 & 31\\
			\hline				  
		\end{tabular}
	\end{center}
	\caption{Effects of knowledge based filters on the process of object identity resolution}
	\label{tab:episode_details}
\end{table}

\newtext{Details of the four runs such as total duration of an episode, number of pick and place tasks (\# pnp tasks), number of objects  at the end of execution in the belief state (\# objs. in bs.) and number of total object hypotheses in the episode (\# of $Hyp$) are presented in Table~\ref{tab:episode_details}. A significant improvement can be noticed in the number of objects in the belief state when filters are turned on.}

\subsection{Knowledge representation}
\label{sec:knowledge_integration}

We choose \knowrob~\cite{tenorth13knowrob} as our knowledge processing 
and inference framework. \knowrob\ is especially designed to be used on robots, can load knowledge, which is stored in the Web Ontology Language OWL~\cite{Bechhofer04OWL}, and can reason about the facts stored using first-order login, implemented in Prolog. 

\begin{figure*}[!t]
  \centering
    \includegraphics[width=0.99\textwidth]{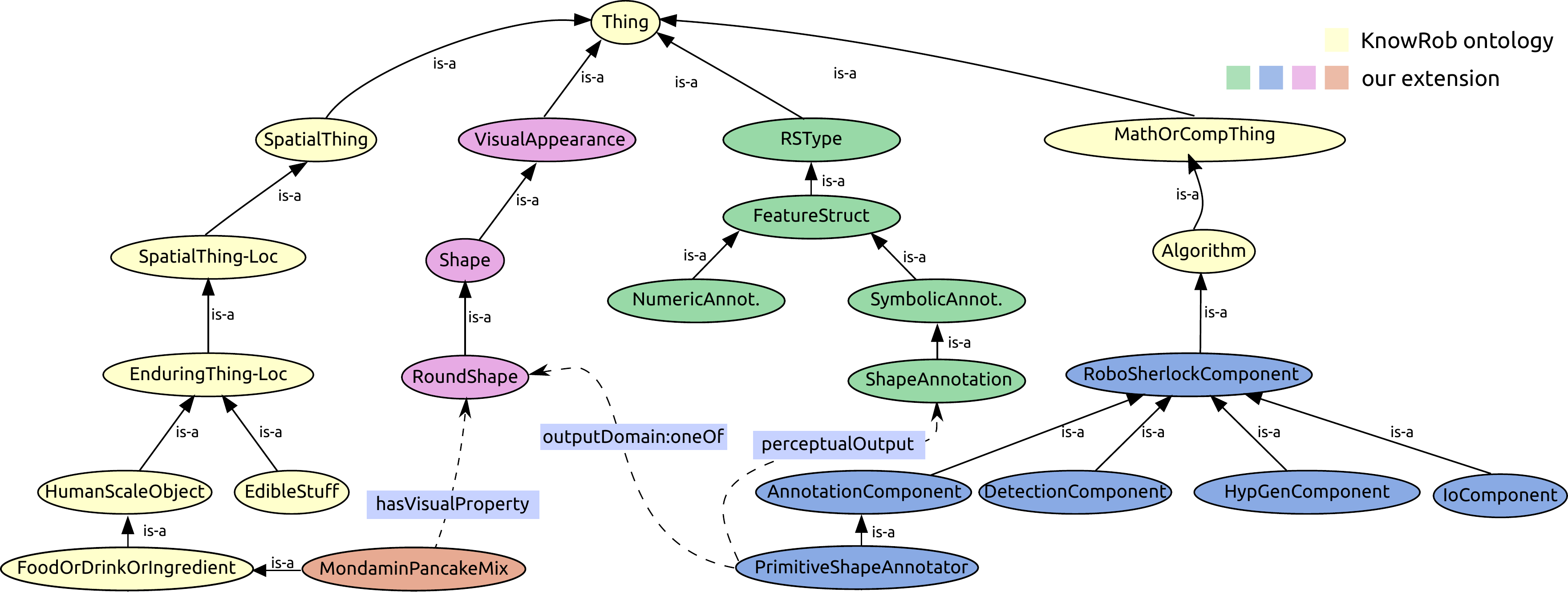}
  \caption{The extended ontology, that is used to reason about perceptual capabilities and generate task- object-dependent perception pipelines}
  \label{fig:ontology}
\end{figure*}

The upper ontology in \knowrob is based on~\citet{opencyc} and is extended with concepts that are relevant to the application domain of robots performing everyday manipulation tasks. To facilitate reasoning about objects of daily use, their visual appearance and the perception capabilities of \robosherlock\ that can detect these visual cues, we extend the available ontology with concepts that represent these. Figure~\ref{fig:ontology} shows parts of the extended ontology, where  different types of classes are color-coded, based on what they describe. By using the already existing ontology as the backbone of our knowledge representation we can reuse existing classes and also ease integration. Namely, we introduce \texttt{RoboSherlockComponent} as a subclass of the already existing \texttt{Algorithm} class, \texttt{VisualAppearance} as a new top-level class and classes for several objects of daily use and a new top level concept for the type system, \texttt{RSType}.

While the main categories in the taxonomy are designed by hand, sub-classes of 
the \texttt{RoboSherlockComponent} and \texttt{RSType} are 
automatically generated, based on definitions in \robosherlock. This is 
possible due to the fact that in a UIM system each implemented PAE has a 
meta description in XML markup. The meta description contains information about 
the type of the expert, its input requirements and its output produced, in 
terms of the type system. Based on these meta descriptions 
we generate the OWL ontology of the perceptual capabilities implemented. 
Finally, the terms presented in the metalanguage are mapped to the entities in the ontology.

The currently implemented perceptual capabilities range from simple image or point cloud segmentation (e.g. plane segmentation) used for hypotheses generation to more complex model-based detectors, deep learning approaches or web-based annotators. These functionalities are referred to as \textit{capabilities} of the perception framework, represented in the ontology as subclasses of \textit{RoboSherlockComponent}. A detailed description of the specific algorithms implemented in \robosherlock\ and used in our experiments can be found in Section~\ref{sec:implementation}. 


As an example of a generated OWL class we present the PrimitiveShapeAnnotator class:

\begin{lstlisting}[style=owl]
Class : PrimitiveShapeAnnotator
  SubClassOf : AnnotationComponent
    dependsOnCapability some Perc3DDepthCapability
    perceptualInputRequired some RsPclNormalsCloud
    perceptualInputRequired some RsAnnotationPlane
    perceptualInputRequired some RsSceneCluster
    perceptualOutput some ShapeAnnotation
    outputDomain: Box, Round, Flat
\end{lstlisting}

\noindent This defines that the annotator requires the normals of the input point cloud, a supporting plane and an existing 3D cluster, and it produces a \texttt{ShapeAnnotation} defined through the \texttt{perceptualOutput} object property. The \texttt{dependsOnCapability} property encodes the fact that the robotic agent needs to be able to perceive depth information in order to use this module. We can optionaly define an \texttt{outputDomain} property that specifies the range of values the annotator can produce.

The children of the \texttt{VisualAppearance} class are used to describe objects through their properties. Using these descriptions, an object class can be described as:

\begin{lstlisting}[style=owl]
Class : MondaminPancakeMix
  SubClassOf: FoodOrDrinkOrIngredient
    hasVisualProperty some MediumSize
    hasVisualProperty some MondaminLogo
    hasVisualProperty some YellowColor
\end{lstlisting}

Properties of objects in the ontology being the same as the output domains of expert algorithms (Figure~\ref{fig:ontology} dashed arrows) makes reasoning about which perception expert to run for different descriptions of objects possible. At the moment properties of the objects found in the knowledge-base are manually inserted, but parts of these could also be learned as we have previously shown in~\cite{icra14ensmln}.

%

Based on the presented ontology of objects and the perceptual capabilities of \robosherlock\ we define Prolog rules to reason over the relation of these. Some of the rules are designed to be generic, such as reasoning about which PAEs result in certain annotations, others can be more specific and tailored to a task. We present the more prominent rules defined, and show their execution. 


\subsection{Pipeline planning and modification}
\label{sec:pipeline-gen}


A perception pipeline in \robosherlock\ is an ordered list of PAEs that are executed  in order to hypothesize and annotate the raw data. In UIM the sequence of PAEs in an aggregate analysis engine (referred to as a perception pipeline from here on), is controllable through a so called \textit{flow-controller}. A flow controller is responsible for the \textit{flow} of the CAS through the AEs. In our current system flow controllers are the implementation of the AAE executors from Figure~\ref{fig:rs_conceptual_view} 

The sequence of PAEs in \robosherlock\ can either be predefined (e.g. as in the continuous execution component) or generated based on the task description. The first set of rules that we define are tasked with inferring which component of \robosherlock\ needs to be run in order to answer a given perception task. This set of rules allow the reconfiguration of the processing pipeline, inferring missing experts and planning their execution sequence. These rules take as an input either the name of an object defined in the knowledge-base or a list of visual characteristics in the form of predicates. The logical rule defined for this in \knowrob\ is:

\begin{lstlisting}[style=prolog]
build_pipeline_for_object(Obj,Pipeline):-
  predicates_for_object(Obj,ListOfPreds),
  enum_annotator_sets_for_pred(ListOfPreds,
		      Annotators),
  build_pipeline(Annotators,Pipeline).  
\end{lstlisting}

\noindent Here, we start with retrieving the perceptual description of the queried object as a list of attributes, then search for the set of annotators that have the desired outputs based on their \texttt{perceptualOutput} object property. Last, we deduce the order of the pipeline using the \textit{build\_pipeline} rule. Simplifying the rule by substituting the object with the list of attributes we can skip the retrieval of these and search for the PAEs directly. For both cases the \textit{build\_pipeline} rule is the same, and is detailed in Algorithm~\ref{alg:algorithm}.

\begin{algorithm}[t]
 \small
 \KwData{robotCaps, $\mathcal{Q}$}
 \KwResult{$AAE^I$  - sequence of PAEs}
 \BlankLine
  $AAE^I$ $\leftarrow$ getExpertsFromContinuousExec()\;
  \uIf {\textbf{not} fulfills($AAE^I$, $\mathcal{Q}$)}{
  $AAE^I$ $\leftarrow$ extendExpertsForQuery($Q$, robotCaps)\;
  }
	//\textit{initialize \textit{$AAE^I$} with PAEs for the given query and robot 
capabilities}\;
  \For{i $\in I $}{
	  prec-experts $\leftarrow$ preconditionsOfExpert($AAE^I(i)$)\;
		\If{ $\exists$ prec-experts $\land$ robotMissesCap(prec-experts, robotCaps)}{
			\Return{An Error}
		}
		\If{ $\exists$ prec-expergs}
			{prec-experts $\leftarrow$ prec-experts $\cup AAE^I$ 
		}
	}

	\While{not completePipeline($AAE^I$)}{
	  $AAE^I$ $\leftarrow$ $AAE^I$ $\cup$ getMissingExperts($AAE^I$)\;
	}
$AAE^I$ $\leftarrow$ sortByDependencies($AAE^I$)\;
\BlankLine
 \caption{Generation of the sequence of PAEs for a query}
 \label{alg:algorithm}
\end{algorithm}

The rule checks if there are any requested types that can't be handled by the continuous execution component and then verifies if the current list of PAEs fulfill the tasks requested by the query (i.e. does the continuous pipeline assert all attributes asked for in the query). If it does not, we calculate the missing annotators beforehand, add them to the list of PAEs to be executed, sort them based on their preconditions between each other and bring them in the correct execution order.

The \texttt{robotMissesCap} rule illustrates how the reasoning 
capabilities about robot hardware and the perceptual capabilities of 
\robosherlock\ are integrated. The capability reasoning uses the SRDL\footnote{semantic robot description language} of a robot, which includes information about the sensors. Consider a PR2 robot we can now ask the system, which perceptual capabilities does the robot possess:

\begin{lstlisting}[style=prolog]
?- perceptual_capabilities_on_robot(Cap,
pr2:'PR2Robot1').
Cap = rs_comp:'Perceive3DDepthCapability' ;
Cap = rs_comp:'PerceiveColorCapability' ;
false.
\end{lstlisting}

\noindent and if a certain \robosherlock\ expert can be run on it: 

\begin{lstlisting}[style=prolog]
?- action_feasible_on_robot(
rs_comp:'Cluster3DGeometryAnnotator', 
pr2:'PR2Robot1').
true.
\end{lstlisting}

When planning a new perception pipeline the above rule is used to determine whether an expert can run or not on the current hardware.

\subsection{Reasoning about scenes and task context}
\label{sec:scene_task}

The use of \knowrob\ concepts and predicates ensures that we can make use of a large body of knowledge about objects, their properties and also robot descriptions. All of the \robosherlock\ component classes in the ontology have the object property \texttt{dependsOnCapability}, which creates a link between perception capabilities of the robotic agent and those of the perception framework. Using existing rules, we can check if our robotic agent has the necessary sensory capabilities to run some expert.

\begin{lstlisting}[style=prolog]
expert_feasible_on_robot(Expert, Robot) :-
   missing_cap_for_action(Expert, Robot, _),
   missing_comp_for_action(Expert, Robot, _).
\end{lstlisting}

We further extend the pipeline planning rules, to enable queries for a broader range of terms, using background knowledge about the objects. Each object in the knowledge-base can have several super-classes, for example a ketchup bottle is a child of the \texttt{Bottle} class as well as the \texttt{FoodOrDrinkOrIngredient} class. The first extension facilitates the recognition of all objects in a scene that are of some certain type. This allows the system to retrieve objects in a scene that are of type food, drink, ingredient or that are bottles or perishable items.

An example rule for such a query is:
 
\begin{lstlisting}[style=prolog]
build_pipeline_for_subclass
		(Obj,Pipeline,ChildObj):-
  owl_subclass_of(Sub, Superclass),
  build_single_pipeline_for_object(ChildObj,
                              Pipeline).
\end{lstlisting}

Replacing the search for subclasses in the above rule with the search for object properties of a class we can further extend it to search and generate perception pipelines for objects that possess a certain property, may that be a visual one (color, texture etc.) or an affordance (has a lid, is graspable etc.). Through the combination of the object representations and the extensions to the pipeline generation rules, the perception system is able to answer queries based on descriptions that are not necessarily related to visual appearances.

Since the results of a perception query are asserted into the knowledge base, it is possible to inspect them and formulate even more complex queries that require further reasoning about the objects properties. In most cases these queries fall in the case of \textit{inspect}ing an object in the meta language. From the name of a detected object we can infer certain properties that can inform the system to run a different set of experts to further analyze the object. For example if the detected object is a container and the detected shape for it was cylindrical the examine pipeline will contain a cylinder matcher to estimate its volume.

\begin{lstlisting}[style=prolog]
detect_volume(Obj,P):-
  owl_subclass_of(Obj,knowrob:'Container'),
  obj_has_predicate(cylindrical_shape, Obj),
  examine_pipeline_for_object(Obj,P).
\end{lstlisting}

Another example would be to search for objects that are most of the time found in combination with other objects but are harder to separate visually(e.g. cutlery can be found near or on plates, a pancake is on the pancake maker etc.). Once we have the already perceived objects asserted we can use the following rule to build special pipelines that take into consideration the dependence between objects as well.

\begin{lstlisting}[style=prolog]
detect_if_individual_present(Obj,
                    DependentObjectClass,P):-
  owl_individual_of(_, DependentObjectClass),
  build_pipeline_for_object(Obj,P).
\end{lstlisting}

These logical rules and combinations of these pave the way for interpreting scenes with the help of knowledge-based reasoning. Through reasoning we can for example infer that plates can be stacked or cups can be found on plates which in turn can inform the perception framework, to take special care of these situations.

\section{Experimental analysis}
\label{sec:experiments}

Since the contributions are neither individual algorithms nor a monolithic system, but a framework, and since it covers a considerably wider scope than previous work, it is hard to quantitatively assess the quality of the proposed approach. It is also hard to compare it to existing perception systems used in robotics, since it builds on top of these systems, and offers developers the possibility to wrap their framework in \robosherlock.\newtext{Evaluating the sub-systems, such as the ones presented as CAS consumers , would also only show how good or bad that specific sub-systems are and is beyond the scope of the current paper.}

We therefore showcase the capabilities of \robosherlock\ on three tasks of a robotic agent performing different experiments: (1) pick and place of objects for table setting (2) a robot performing pipetting in a chemical laboratory (3) robot operating in a supermarket. All of these scenarios are parts of past or ongoing EU projects where \robosherlock\ was or is the main perception engine.

We evaluate the proposed system by formulating perception queries for each scenario, showing the variety of formulations. We exemplify the use of knowledge based reasoning, demonstrating the broad range of queries the system can handle.

Let us start by considering  a simple query in the proposed metalanguage, where we are interested in all black objects that are round and are located on a counter top:

\begin{small}
	\begin{tabbing}
		(detect (an \= object (shape round) \\
		\> (color black) \\
		\> (location (a \=location ( \\
		\> \>on (a table (category table-top\#3)))))
	\end{tabbing}
\end{small}

The task interpreter will search for the terms of the query and plan for a pipeline that can detect these, taking into consideration the restrictions on output values and input requirements each PAE might have:

\begin{lstlisting}[style=prolog]
?- pipeline_from_predicates([color,shape,
                            location],AAE).
AAE = [
rs_comp:'PlaneAnnotator',
rs_comp:'PointCloudClusterExtractor',
rs_comp:'NormalEstimator',
rs_comp:'PrimitiveShapeAnnotator',
rs_comp:'ClusterColorHistogramCalculator',
rs_comp:'ClusterLocationAnnotator'].
\end{lstlisting}

Even though we queried only for shape, color and location (the last three components in the result), the system is able to deduce the missing components and build a complete perception pipeline containing hypotheses generator, data pre-processors and the annotators. Once this pipeline is executed the result interpreter will find the object hypotheses that correspond to the description, return their unique ID and assert them in the knowledge base. This will have further implications, as we will demonstrate. We now have a detailed look at the perception tasks involved in our three demonstrator scenarios and how they are handled by the framework. 

\begin{figure*}[!t]
	\centering
	\includegraphics[width=0.33\columnwidth]{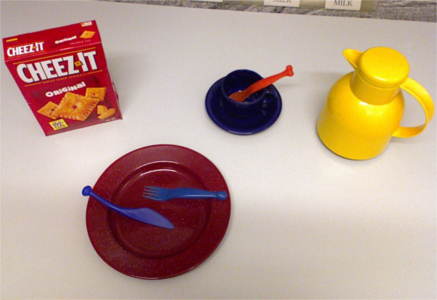}
	\includegraphics[width=0.33\columnwidth]{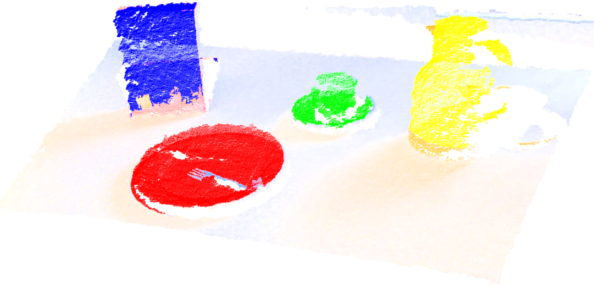}
	\includegraphics[width=0.33\columnwidth]{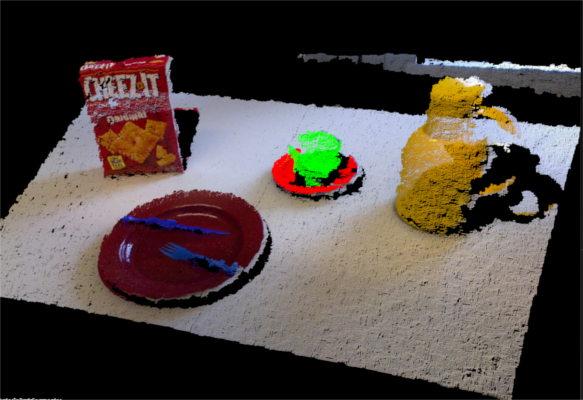}
	\includegraphics[width=0.33\columnwidth]{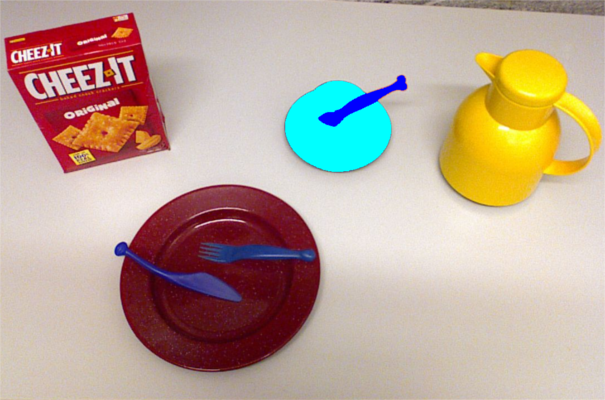}
	\includegraphics[width=0.33\columnwidth]{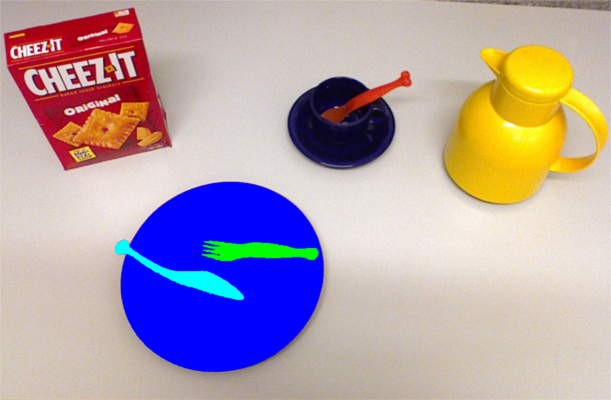}
	\includegraphics[width=0.33\columnwidth]{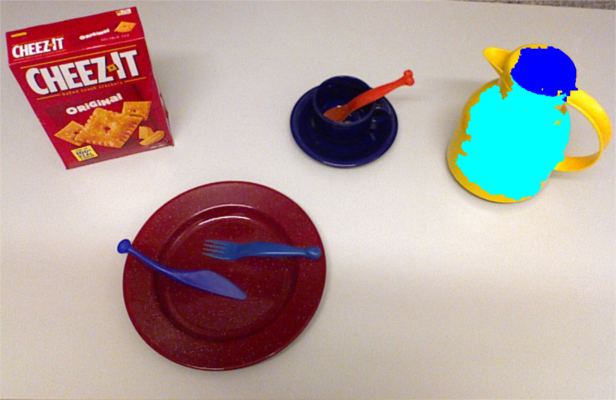}\\
	\vspace{0.5mm}
	\includegraphics[width=0.305\columnwidth]{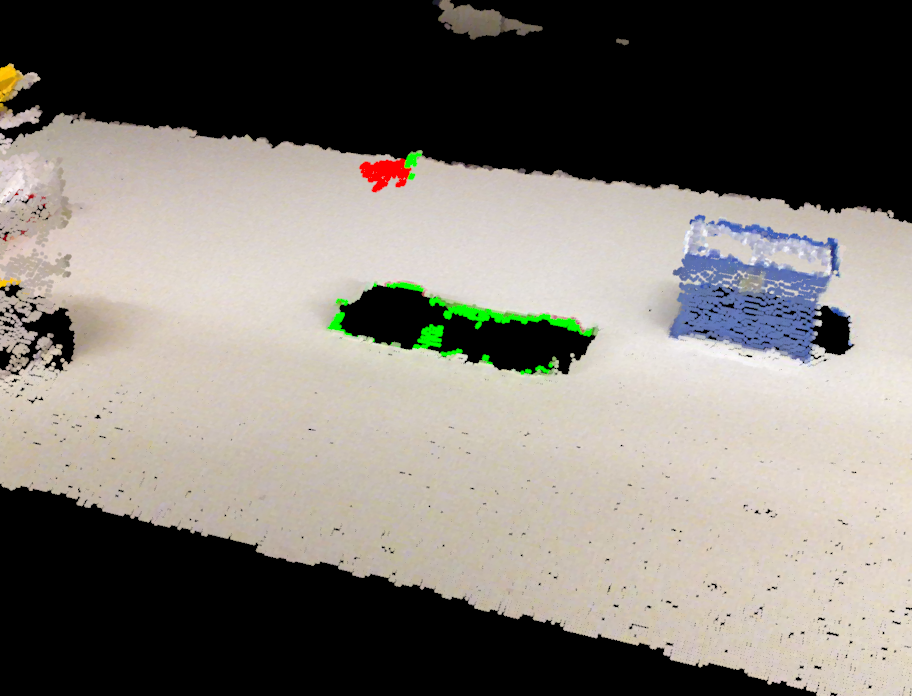}
	\includegraphics[width=0.33\columnwidth]{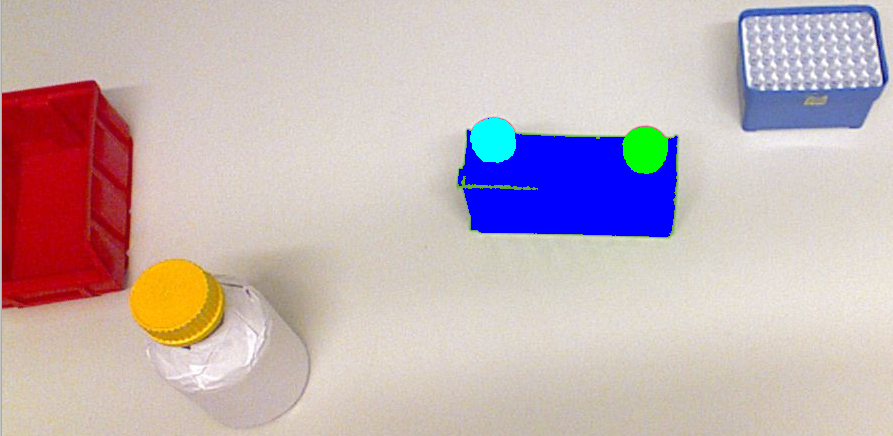}
	\includegraphics[width=0.33\columnwidth]{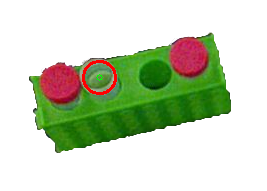}
	\includegraphics[width=0.33\columnwidth]{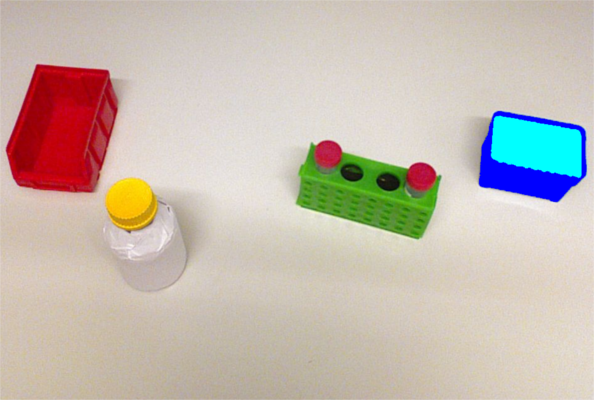}
	\includegraphics[width=0.365\columnwidth]{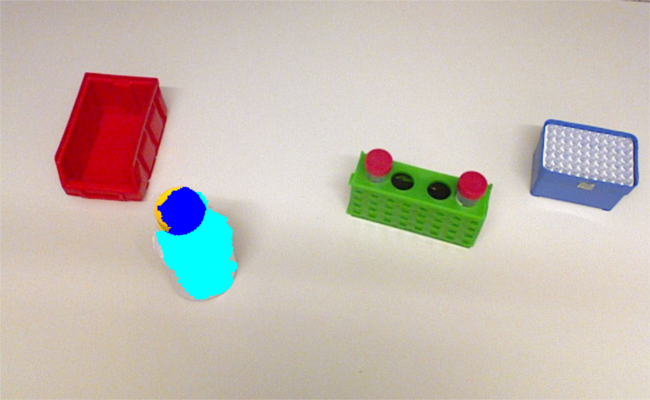}
	\includegraphics[width=0.30\columnwidth]{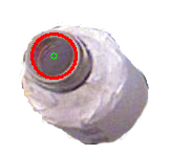}
	\caption{Result of inspection tasks using background knowledge about the 
		asserted objects in a breakfast scene and in 
		a chemical lab scene. \textbf{Top row, left to right}: original image, 
		segmentation results, hypotheses about: a cup on 
		a plane, spoon in a cup, cutlery on plate, the lid of the pitcher. 
		\textbf{Bottom row, left to right}: raw point cloud 
		data of the tubes rack, hypotheses about the tube caps, opening of a 
		tube, hypothesis about the opening of the pipette 
		tips container, lid of a bottle and opening of a bottle}
	\label{fig:hypotheses_inspections}
	\vspace{-3ex}
\end{figure*}

\subsection{Task 1: Setting a table in a kitchen}

\begin{figure}[t!]
	\centering	
	\includegraphics[width=0.49\columnwidth]{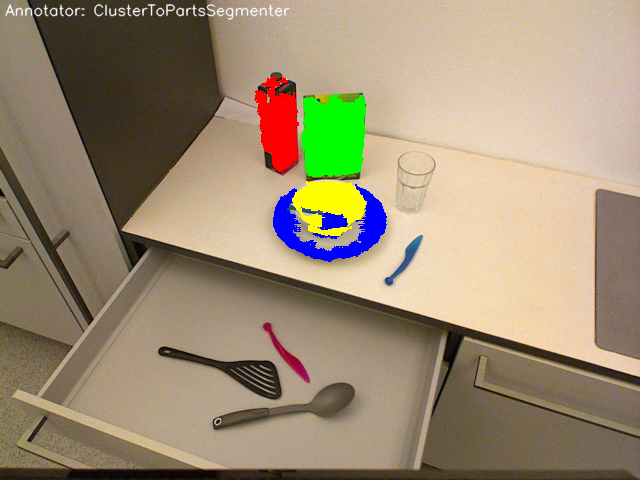}
	\includegraphics[width=0.49\columnwidth]{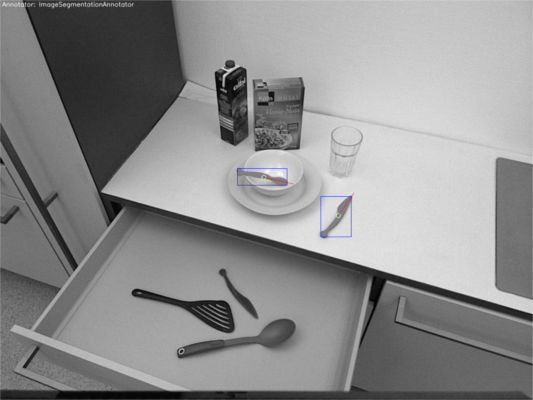}\\
	\includegraphics[width=0.49\columnwidth]{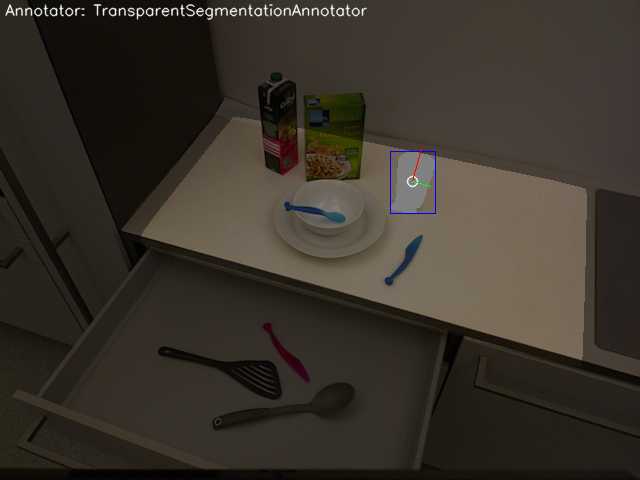}
	\includegraphics[width=0.49\columnwidth]{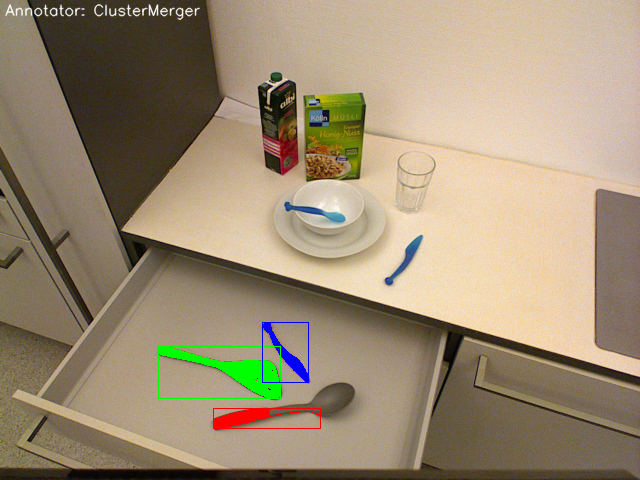}
	\caption{Results of various hypotheses generation algorithms, using the visualization from \robosherlock. Top-left: 3D region growing, Top-right: binary segmentation, Bottom-left: transparent segmentation, Bottom-right: specific semantic region}
	\label{fig:hyp_gen}	
\end{figure}

The first task we analyze is that of a household robot setting and cleaning a table, performing pick and place tasks. Perception tasks will be exemplified on the scene from Figure~\ref{fig:example_scene}. 

In this scenario the continuous component is configured to look for objects of daily use on supporting planes using region growing based on euclidean distance in 3D and 
normal space. Movable objects are not the only ones of interest in this scene. 
If we want to put away the clean objects we might want to open drawers. The 
control system can direct \robosherlock\ to do just that by issuing the query:

\begin{minipage}{0.5\textwidth}
	\centering
	\begin{footnotesize}
		\begin{tabbing}
			(\textbf{d}\=\textbf{etect} ( an object (\\
			\>(type 'Handle')\\
			\>(part-of (an \=object\\
			\>\>( type 'Drawer'))\\
			))))
		\end{tabbing}
	\end{footnotesize}
\end{minipage}
\hspace{0.0cm}
\begin{minipage}{0.50\textwidth}
	\includegraphics[width = 0.5\columnwidth]{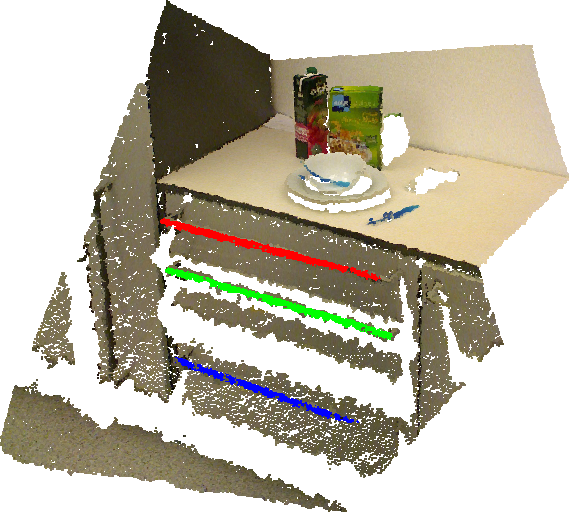}
\end{minipage}

The framework deduces that it needs to switch to a set of PAEs that were specially tailored to find the handles in our kitchen and execute that. This is done by joint reasoning about that the attributes the query is asking for (type, part-of) and what the values of these attributes are ('Handle'). 

The default hypotheses generation only finds a subset of the objects. By querying for specific objects we can task the framework to look for the transparent glass, or the knife. Similarly by specifying which semantic location we want the robot to look at we can find the objects in the drawer. Results of the hypotheses generation are shown in Figure ~\ref{fig:hyp_gen}. As it can be observed, transparent or very flat objects would not be found with the default table top segmentation. 


Before grasping any of the objects, the robot might want to take a closer look at them. There can be multiple reasons for doing so. Either we want to check the state of the object being manipulated (is the plate dirty, is there something in the bowl) or  get a better pose estimate using for example an additional camera mounted on the wrist of the robot. These cases can be handled by issuing inspection queries for objects. 

\begin{tabbing}
(\textbf{i}\=\textbf{nspect} (\#uid :pose,:obj-part)
\end{tabbing}

Because the objects perceived using the detection queries are asserted into the world model, these inspection queries are highly dependent on the background knowledge we have stored about them. If for instance a hypotheses was detected as being an object of type container the system can run over-segmentation algorithms on the hypothesis in order to examine them. Results of queries where we are examining previously asserted objects in the belief state are shown in Figure~\ref{fig:hypotheses_inspections}.

We should also be able to detect all similar objects based on some common property. For instance all cutlery or all perishable items. This can help optimize the sequence of steps it has to take in order to clean the table. Using the query language we can formulate several similar queries that would yield the same result:

\vspace{1ex}
\begin{minipage}{0.5\textwidth}
	\centering
	\begin{footnotesize}
		\begin{tabbing}
			(\textbf{d}\=\textbf{etect} ( an object(\\
			\>(shape box) (color green)))\\
			(\textbf{d}\=\textbf{etect} ( an object(\\
			\>(class 'KnusperHonig'))))\\
			(\textbf{d}\=\textbf{etect} ( an object(\\
			\>(type 'Food')))\\
		\end{tabbing}
	\end{footnotesize}
\end{minipage}
\hspace{0.1cm}
\begin{minipage}{0.50\textwidth}
	\includegraphics[width = 0.5\columnwidth]{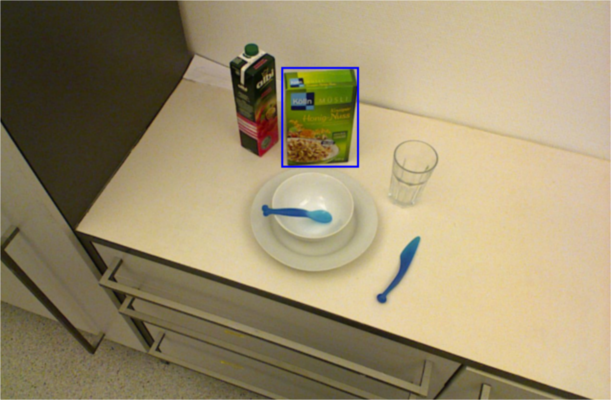}
\end{minipage}
\vspace{1ex}

These queries would all yield the same result, the green cereal box, and illustrate the expressiveness of the query language. Asking for an object of a certain type ('Food' in our example), signals to the system that the results of annotators should be reasoned upon before sending back a response. This way we can query for objects using properties that are not directly perceivable.

\subsection{Task 2: Chemical experiment}
\label{sec:t2_chemlab}

A second task that we look at in is performing parts of a DNA extraction process~\cite{lisca15osd}. Although in this task a closed world assumption holds, the objects used and the variety of relationships between these makes the application highly challenging (transparent objects, objects with (re)movable parts), with buttons etc.). It serves as a great example of how the high level control program can steer perception in the right direction through semantic querying.

\begin{figure}[t]
	\begin{center}
		\includegraphics[width=0.49\columnwidth]{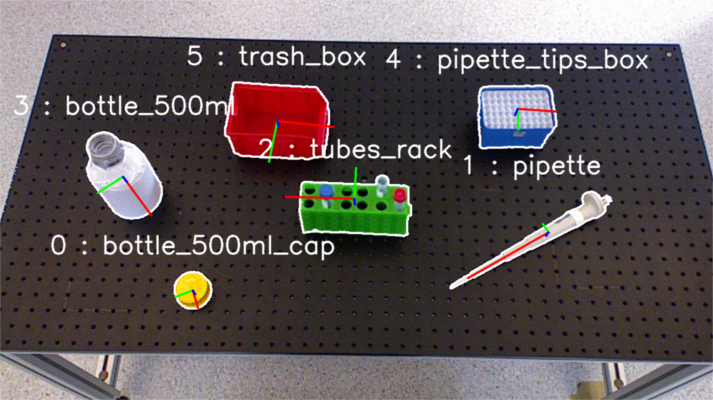}
		\includegraphics[width=0.49\columnwidth]{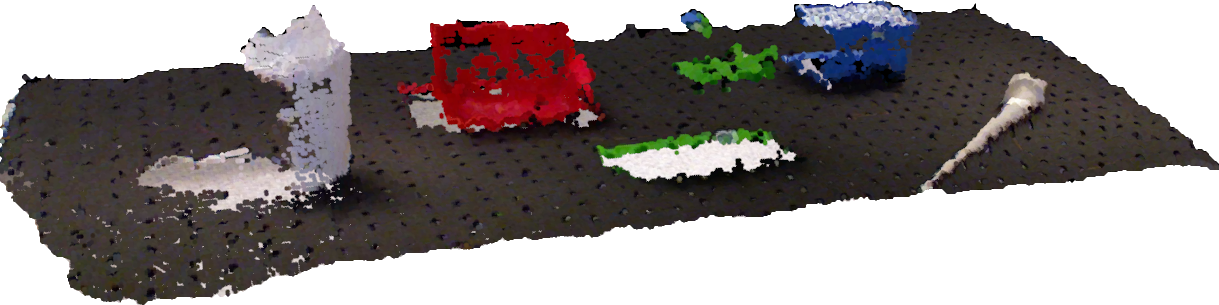}
	\end{center}
	\caption{Pipetting scene as seen by the robot. \textit{Left: RGB, Right:
			PointCloud}}
	\label{fig:pipetting_scene}
\end{figure}

The robots' task is to pick up the pipette, mount a tip on it, get some solution from the bottle and release it into one of the tubes found in the rack. Finally the robot has to release the contaminated pipette tip into the trash box. The challenges for perception here are: (1) not all objects can be perceived using RGB-D sensors (see  Fig.~\ref{fig:pipetting_scene}), hence we use a combination of color segmentation and point cloud clustering and (2) some of the perception tasks needed are based on common sense knowledge: the opening of a container is the top part of an object, tubes are to found in a tube-rack, can be open or closed etc.

Queries like detect the bottle with acid in it, or detect the Ellemayer flask on the heater that can hold 500ml of liquid are possible due to the representation of objects in the knowledge base and the flexible re-planning capabilities of the system.

There is no one unique solution to perceive all of the objects, and their parts,
but having several expert algorithms combined with knowledge processing
significantly increases the success rate of finding the relevant items on the
table. We precisely need to identify where the pipette tip needs to enter the 
bottle, or find the tubes in the rack. Accomplishing this task is possible 
through formulating rules like: 

\begin{sl}
\begin{footnotesize}
  \begin{tabbing}
    fit\=Circle(Obj,Radius) :- \\
    \> category(Obj,'container'), object-part(Obj,Opening),\\
    \> geom-primitive(Obj,'circular'), \\
    \> radius(Opening,Radius),\\
  \end{tabbing}
\end{footnotesize}
\end{sl}
\vspace{-3ex}

\noindent which deduces the radius of the circle that needs to be fit in order to find
the opening on top a container, or the holes in the rack where tubes can be found.

The reasoning mechanisms can enable detection of objects where there is little to no, or very noisy sensory input. For instance the floating small segment (highlighted in red in the bottom left corner of Figure~~\ref{fig:hypotheses_inspections}) signals the system to search for hypotheses in image space, hence we find the tube rack. Asserting this informs the system that it should search for tubes in it, leading to a final detection of the tubes, by applying a specialized expert.

\subsection{Task 3: Object counting in a Supermarket}

\begin{figure}[t!]
	\centering
	\includegraphics[width = 0.99\columnwidth]{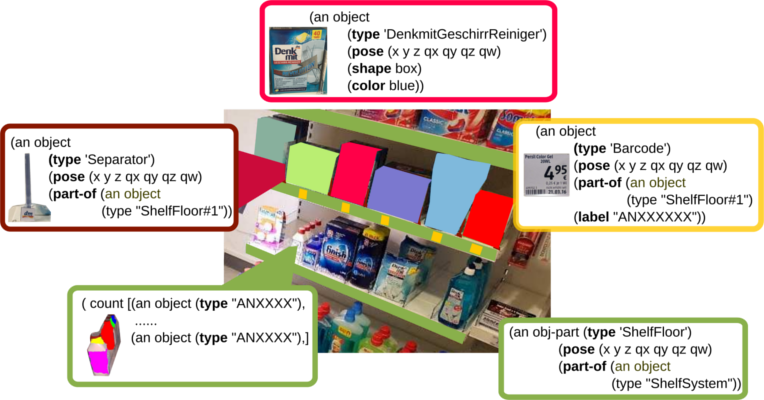}
	\caption{Semantically rich description of a scene from the a retail store}
	\label{fig:example_scene_refills}
\end{figure}

The last task we look at is that of building and maintaining a semantic object map in a retail environment. Figure~\ref{fig:example_scene_refills} shows an example scene from our mock-up store. The perception tasks that are needed in order to successfully manage a semantic map vary greatly. This scenario highlights the need for taskability and adaptability of the framework. 

A store consists of several modular elements that can often change positions. Shelf systems are considered to static objects, but shelf floors, barcodes on the shelf facings, product separators and obviously the products themselves can change location often. The robot is tasked with scanning the shop every night and rebuilding a semantic object store. The perception system of such a robot is responsible for detecting all of the movable parts of the shop. It is evident that the detection of these parts can only be addressed with very different algorithmic solutions and that if we want a truly autonomous system, taskability is a key concept. Detecting shelf floors for instance can be handled by fitting lines using RANSAC or Hough transform, given the right conditions are met (camera and robot position, viewing angles etc.). On the other hand a volumetric counting algorithm using background knowledge about the objects can be used to estimate the number of objects of a certain type. Running any of these algorithms in the wrong context would yield erroneous or no results at all. 

Some of the perception tasks in this scenario are what we call continuous 
tasks, and  need continuous observations in order to correctly hypothesize about object location. We can use the compound querying capabilities of \robosherlock\ to 
state such tasks. For instance scanning for the shelf floors, barcodes on a 
shelf facing, or the separators can be formulated as:

\begin{minipage}{0.5\textwidth}
\centering
	\begin{footnotesize}
		\begin{tabbing}
		(\textbf{scan} \= (for \= object\\
		\>(detect ( an object (\\
			\>\>(type 'Shelf')\\
			\>\>(command 'start')\\
		))))
		\end{tabbing}
	\end{footnotesize}
\end{minipage}
\hspace{0.6cm}
\begin{minipage}{0.5\textwidth}
	\includegraphics[width = 0.5\columnwidth]{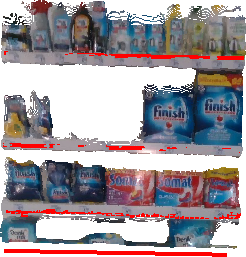}
\end{minipage}

This command will have as direct effect the exchange of the continuous 
component with a perception pipeline that contains the expert for shelf 
scanning. Through the \textit{command} attribute the control program running on the robot can control the continuous execution.  Using the encapsulated detection query we can specify what kind of object do we want the system to scan for. 

A similar compound query can be used to count objects of a certain type:
 
	\begin{minipage}{0.5\textwidth}
		\centering
		\begin{footnotesize}
			\begin{tabbing}
				(\textbf{co}\=\textbf{unt} (object\\
				\>(d\=etect  ( an object (\\
				\>\>(type 'ANXXXX')\\
				\>\>(pose (x y z qx qy qz qw)) \\
				\>\>(width 0.05)\\
				)))))
			\end{tabbing}
		\end{footnotesize}
	\end{minipage}
	\hspace{0.0cm}
	\begin{minipage}{0.5\textwidth}
		\includegraphics[width = 0.5\columnwidth]{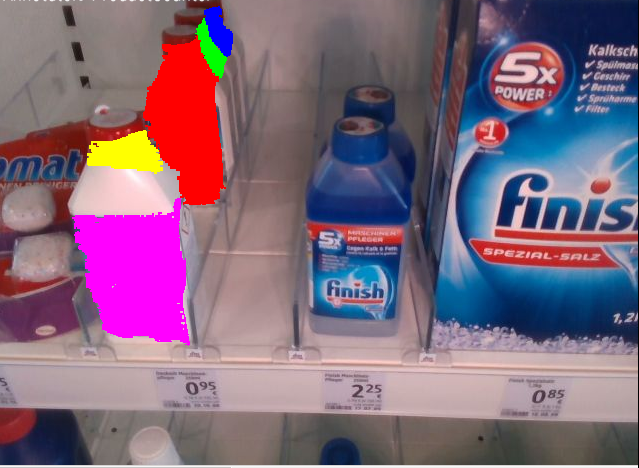}
	\end{minipage}%
\vspace{1ex}

The entire retail scenario is highly knowledge intensive. Stores usually have large databases where information about objects useful to the perception system is kept. This knowledge can help in greatly simplifying the requirements of the perception system. Information such as a product image can help verify if the objects we are perceiving are placed at the correct location or information about the size of an object can help filter the raw data as well as facilitate volumetric counting. 

\paragraph*{}
The three scenarios shown here are only a subset of the tasks \robosherlock\ 
has been deployed in. The flexibility and re-usability of perception experts 
make the framework applicable in very versatile task. We now look at how 
our work relates to the existing body of the previous works in robot perception and how it can be seen as complementary to these. 

\section{Related Work}
\label{sec:related_work}

There are a lot of works oriented at creating perception libraries which are a collection of task specific algorithms, e.g. PCL, OpenCV and the V4R library~\footnote{https://www.acin.tuwien.ac.at/vision-for-robotics/software-tools/v4r-library/}. The perception tasks demanded by the scenarios we just described go substantially beyond what is supported by current perception libraries and frameworks. What we mean is that these frameworks, while excellent from the perspective of algorithmic performance, do not encode the knowledge to automatically address higher level perception tasks.

Frameworks, mostly based on middle-ware like ROS, such as SMACH~\cite{bohren11butlers}) or REIN~\cite{rein11ICRA} have targeted the ease of program development but the problems of boosting perception performance through more powerful method combination has received surprisingly little attention. 
An early example of a robotic perception system was described by Okada \textit{et al.}~\cite{okada2007iros}, where a particle filter based integration of multiple detectors and views was achieved. The probabilistic fusion of different results corresponds to a simple rule ensemble, i.e. one that is not trainable. Similar methods have been employed for semantic mapping approaches \cite{Pronobis10IJRR,mozos11furniture}. More recent work on general frameworks  for robotic agents \cite{wyatt10tamd,scheutz06aai} approach the problem in a much broader sense, thus do not address the specific needs of robot perception.

A lot of existing perception systems usually consider the case where a database of trained object is used to match it with sensor data. Even more, many systems focus on individual algorithms that work on objects with specific characteristics, e.g.\ point features for 3D opaque  objects~\cite{Aldoma2012RAM}, visual keypoint descriptor based systems like MOPED~\cite{Collet2011} for textured or \cite{Lysenkov-RSS-12} for translucent objects. \robosherlock\ is capable  of incorporating all of these different frameworks and combine their results meaningfully to build on the strengths and mitigate the weaknesses of the individual methods.

In the field of knowledge-based vision there have been many systems proposed, that are mainly focused on interpreting the raw sensor data with knowledge on different levels ~\cite{hudelot_towards_2005}~\cite{hartz_learning_2007}. ~\citet{hudelot_towards_2005} created an ontology with signal features used in image processing and included knowledge about processing algorithms, that produces this data. However, the work lacks a unified knowledge representation, different sources of information being stored and processed in separate knowledge bases. This makes reasoning over the whole of the represented knowledge tedious. A lot of work has also been done on developing different visual primitives (\cite{town_ontological_2006}, \cite{hudelot_towards_2005}), which are domain-independent descriptions of visual data. 

Perhaps the most relevant statement from recent literature that addresses knowledge-based perception is formulated by Fiorini in ~\cite{fiorini_review_2010}. In this review about knowledge-based computer vision the author argues that the greatest challenge lies in the appropriate representation of visual knowledge needed for perception tasks, emphasizing the need for standardization. Although the representations presented in our work should not be  considered as standards, we offer a unique solution where perceptual capabilities and visual knowledge are tightly coupled together.  

~\citet{hager2014keynote} identifies two main future challenges of robot sensing.  The design of modular sensing, organized around ontologies and information structure rather than the data itself and the realization of context-relevant question answering systems through ``information APIs``.

In robotics one of the earlier predecessors of \robosherlock\ that aimed at run-time reconfigurability is RECIPE~\cite{arb98sys}. COP-MAN~\cite{IAS09CoPMan} is another interesting approach for bringing knowledge processing and robotic perceptual capabilities closer together. There has also been a lot of work towards knowledge representation in robotics~\cite{tenorth13knowrob} and combination of knowledge processing and perception~\cite{iros10kcopman}. In the later work Dejan et.al. present a pure SWI-Prolog based interface for perception. The approach is limited in the types of queries it can answer and  would also not scale with the number of perception routines available. All of these previous works address only parts of the problem and the experience gathered through developing them formed the basis the current framework.

\newtext{A recent highly integrated robotic learning system is presented by ~\citet{skocajInteractive}, focused on interactively learning to associate linguistic and visual percepts by acquiring abstract knowledge of  generic categories like object color, shape and type. The authors start with a comprehensive presentation of the fields development, and present a statistical approach for identifying lacking information. To fill this knowledge-gap a human operator can be queried and perceptual tasks be performed automatically, or at the direction of the human. The presented system is highly complementary to \robosherlock, as it focuses on building up a model database, learning what different shape, color and object type categories mean, and how to continuously update these beliefs using sensory data and interactions with humans. For these tasks \robosherlock relies on an existing representation of world and in this article we focus only on querying such common sense knowledge, instead of continuously learning it.}

\newtext{A related approach for actively generating pipelines to solve perceptual tasks was described by ~\citet{Sridharan10planningtToSee}, based on an efficient hierarchical POMDP formulation. The authors also focus on color, shape and type categories as ~\citet{skocajInteractive}, and answer queries on object location, existence and identity, i.e. they also consider clutter segmentation by including a ROI splitting operator (which would be a hierarchical hypothesis generator in RS). The ROI splitting is a specific challenge for the POMDP, but overall the operators are simple, static and few. In contrast, \robosherlock\ generates pipelines that are considerably more complex, and utilize background knowledge, but currently lack the probabilistic sequential decision making during executions.}

In recent years the research in the vision community has been dominated by deep learning. Initially only focusing on recognizing specific object instances and classes recently the focus has shifted to recognizing and detecting objects~\cite{convDetectorsSurvey} ans estimating their pose~\cite{DBLP:journals/corr/abs-1803-08103}. One of the first well performing algorithms that tackled object detection as a neural network regression problem is called YOLO and was presented in \citep{redmon2016yolo9000}. Earlier algorithms usually used two separate steps for detection and classification, e.g. \citep{girshick14CVPR}. The main advantages over those methods are the much faster processing speeds and the simple end-to-end training of the algorithm.

\citet{liu2016ssd} proposed a method called SSD to extend any existing image classification neural network to support predicting of object bounding boxes. To that end the existing network is truncated before the classification layers and an auxiliary structure, consisting of several convolutional layers, is added. \citet{lin2017focal} recently introduced RetinaNet, with a proposed improvement to the loss function of object detectors, and obtained an upper envelope on existing methods when tuned for different inference time vs accuracy. 

While results are impressive and in some cases recognition rates are comparable with human performance  these approaches are rather complementary to the \robosherlock\ framework, and can act as highly capable experts that hypothesize or annotate data. 

Using commercial systems like the one offered by Cognex~\footnote{https://www.cognex.com/products/leading-technology/deep-learning-based-image-analysis}, VIDI\footnote{http://imaging.market/vidi-green-object-scene-classification-10000337} all you need to do is provide a sufficiently large training set and you have a perception expert that can solve the specific task. The idea behind \robosherlock\ is to build perception pipelines from the individual experts, not having to retrain networks for every single perception task that needs solving.

One of the exciting areas of research in deep learning, that relates well to our work is that of visual query answering and description generation based on images. According to the survey of the image to caption field by \citet{bernardi2016automatic}, our work could be categorized as a \emph{direct generation model}, whereby detections are made and then used by a following description generation step. However the detection is guided by the task and by the a priori cues it receives from the high-level planning system. Caption generation can be performed using RNNs, for example by mapping sentences and images to a latent representation where they are related \cite{Socher2014GroundedCS}, and by applying LSTM or other architectures for further improvement \cite{Donahue2017PAMI}. However, most of them need image-sentence pairs for training, and the substantial time for retraining a network every time makes them so far not applicable.


\section{Discussion}
\label{sec:discussion}

We have presented a cognitive perception framework that treats perception as a query answering problem and can generate context-specific perception plans and adapt to the demands of manipulation tasks. Through reasoning about background knowledge the system is able to answer a wide variety of possible questions going beyond what is directly perceivable. One of the key questions is how can this scale towards a large amount of objects. Our aim is to create a hybrid system, that uses knowledge processing in situations where it would be difficult for general purpose algorithms to produce reliable results. This means, that scaling is dependent on having experts in \robosherlock\ that can handle large amounts of object classes reliably. The same holds for handling very cluttered scenes. \robosherlock\ already integrates many state of the art algorithms for hypothesizing about objects in cluttered, occluded environments (e.g. SimTrack, or MOPED). Using background knowledge and reasoning to enhance the recognition rates for such scenes.

\subsection*{Extensions}

The components of \robosherlock\ as presented in this work constitute only the 
core functionalities. As closing thoughts we would like to offer an overview of 
the extensions that have been made possible by these core functionalities.  
These extensions further extend the cognitive capabilities of the framework. 
It is essential to collect the memories of a robotic agent during task 
execution in order to improve over time. The perceptual episodic memories that \robosherlock\ stores have been presented in~\cite{balintbe17icar} and improving the results of individual expert based on these has been prototyped. In \cite{worch2015perceptionHRI} we presented an extension that allows for detecting humans and objects at the same time, while simultaneously reasoning about the possible actions that are being performed. Retrospectively allowing to investigate what action a human performed, how it did it and what its intention was. A system for pervasively tracking and solving identities of objects was proposed in ~\cite{Wiedemeyer15pervasive} on which we have recently built an amortized perception component that spreads the cost of answering queries throughout the execution of the robot task. More recently we have investigated how through the \robosherlock\ framework one can enable the agents to ``dream``~\cite{balint18iros}, and using state of the art gaming engines generate variations of a task and learn new perception models. All of these extensions look at robot perception from the perspective of a robot performing tasks, which would not have been possible without the core framework presented here.


\begin{thebibliography}{59}
\providecommand{\natexlab}[1]{#1}
\providecommand{\url}[1]{\texttt{#1}}
\providecommand{\urlprefix}{URL }
\expandafter\ifx\csname urlstyle\endcsname\relax
  \providecommand{\doi}[1]{DOI:\discretionary{}{}{}#1}\else
  \providecommand{\doi}{DOI:\discretionary{}{}{}\begingroup
  \urlstyle{rm}\Url}\fi

\bibitem[{Aldoma et~al.(2012)Aldoma, Marton, Tombari, Wohlkinger, Potthast,
  Zeisl, Rusu, Gedikli and Vincze}]{Aldoma2012RAM}
Aldoma A, Marton ZC, Tombari F, Wohlkinger W, Potthast C, Zeisl B, Rusu RB,
  Gedikli S and Vincze M (2012) {Tutorial: Point Cloud Library --
  Three-Dimensional Object Recognition and 6 DoF Pose Estimation}.
\newblock \emph{Robotics \& Automation Magazine} 19(3): 80--91.

\bibitem[{Arbuckle and Beetz(1998)}]{arb98sys}
Arbuckle T and Beetz M (1998) {RECIPE} - a system for building extensible,
  run-time configurable, image processing systems.
\newblock In: \emph{Proceedings of Computer Vision and Mobile Robotics (CVMR)
  Workshop}. pp. 91--98.

\bibitem[{Balint-Benczedi and Beetz(2018)}]{balint18iros}
Balint-Benczedi F and Beetz M (2018) Variations on a theme:'it's a poor sort of
  memory that only works backwards'.
\newblock In: \emph{International Conference on Intelligent Robots and
  Systems}. IEEE.
\newblock \doi{10.1109/IROS.2018.8594001}.

\bibitem[{Balint-Benczedi and Beetz(2019)}]{balintbe19amortized}
Balint-Benczedi F and Beetz M (2019) Amortized object and scene perception for
  long-term robot manipulation .

\bibitem[{Balint-Benczedi et~al.(2016)Balint-Benczedi, Mania and
  Beetz}]{balintbe16task}
Balint-Benczedi F, Mania P and Beetz M (2016) Scaling perception towards
  autonomous object manipulation --- in knowledge lies the power.
\newblock In: \emph{International Conference on Robotics and Automation
  (ICRA)}. Stockholm, Sweden.
\newblock
  \urlprefix\url{http://ieeexplore.ieee.org/stamp/stamp.jsp?arnumber=7487801}.

\bibitem[{Balint-Benczedi et~al.(2017)Balint-Benczedi, Marton, Durner and
  Beetz}]{balintbe17icar}
Balint-Benczedi F, Marton ZC, Durner M and Beetz M (2017) Storing and
  retrieving perceptual episodic memories for long-term manipulation tasks.
\newblock In: \emph{Proceedings of the 2017 IEEE International Conference on
  Advanced Robotics (ICAR)}. Hong-Kong, China.
\newblock Finalist for Best Paper Award.

\bibitem[{Bechhofer et~al.(2004)Bechhofer, van Harmelen, Hendler, Horrocks,
  McGuinness, Patel-Schneider and Stein}]{Bechhofer04OWL}
Bechhofer S, van Harmelen F, Hendler J, Horrocks I, McGuinness DL,
  Patel-Schneider PF and Stein L (2004) {OWL Web Ontology Language Reference}.
\newblock \urlprefix\url{http://www.w3.org/TR/owl-ref/}.
\newblock W3C Recommendation.

\bibitem[{Beetz et~al.(2015)Beetz, Balint-Benczedi, Blodow, Nyga, Wiedemeyer
  and Marton}]{beetz15robosherlock}
Beetz M, Balint-Benczedi F, Blodow N, Nyga D, Wiedemeyer T and Marton ZC (2015)
  {RoboSherlock: Unstructured Information Processing for Robot Perception}.
\newblock In: \emph{IEEE International Conference on Robotics and Automation
  (ICRA)}. Seattle, Washington, USA.
\newblock
  \urlprefix\url{https://ieeexplore.ieee.org/stamp/stamp.jsp?tp=&arnumber=7139395}.
\newblock Best Service Robotics Paper Award.

\bibitem[{Beetz et~al.(2009)Beetz, Blodow, Klank, Marton, Pangercic and
  Rusu}]{IAS09CoPMan}
Beetz M, Blodow N, Klank U, Marton ZC, Pangercic D and Rusu RB (2009) {CoP-Man
  -- Perception for Mobile Pick-and-Place in Human Living Environments}.
\newblock In: \emph{Proceedings of the 22nd IEEE/RSJ International Conference
  on Intelligent Robots and Systems (IROS) Workshop on Semantic Perception for
  Mobile Manipulation}. St. Louis, MO, USA.
\newblock Invited paper.

\bibitem[{Bernardi et~al.(2016)Bernardi, Cakici, Elliott, Erdem, Erdem,
  Ikizler-Cinbis, Keller, Muscat, Plank et~al.}]{bernardi2016automatic}
Bernardi R, Cakici R, Elliott D, Erdem A, Erdem E, Ikizler-Cinbis N, Keller F,
  Muscat A, Plank B et~al. (2016) Automatic description generation from images:
  A survey of models, datasets, and evaluation measures. .

\bibitem[{Blodow(2014)}]{blodow14phd}
Blodow N (2014) \emph{{Managing Belief States for Service Robots: Dynamic Scene
  Perception and Spatio-temporal Memory}}.
\newblock PhD Thesis, Intelligent Autonomous Systems Group, Department of
  Informatics, Technische Universit\"at M\"unchen.

\bibitem[{Bohren et~al.(2011)Bohren, Rusu, Jones, Marder-Eppstein, Pantofaru,
  Wise, Mosenlechner, Meeussen and Holzer}]{bohren11butlers}
Bohren J, Rusu RB, Jones EG, Marder-Eppstein E, Pantofaru C, Wise M,
  Mosenlechner L, Meeussen W and Holzer S (2011) Towards autonomous robotic
  butlers: Lessons learned with the pr2.
\newblock In: \emph{ICRA}. Shanghai, China.

\bibitem[{Bradski(2000)}]{opencv_library}
Bradski G (2000) {The OpenCV Library}.
\newblock \emph{Dr. Dobb's Journal of Software Tools} .

\bibitem[{Codd(1970)}]{codd1970relational}
Codd EF (1970) A relational model of data for large shared data banks.
\newblock \emph{Commun. ACM} 13(6): 377--387.
\newblock \doi{10.1145/362384.362685}.
\newblock \urlprefix\url{http://doi.acm.org/10.1145/362384.362685}.

\bibitem[{{Collet Romea} et~al.(2011){Collet Romea}, {Martinez Torres} and
  Srinivasa}]{Collet2011}
{Collet Romea} A, {Martinez Torres} M and Srinivasa S (2011) {The MOPED
  framework: Object recognition and pose estimation for manipulation}.
\newblock \emph{International Journal of Robotics Research} 30(10): 1284 --
  1306.

\bibitem[{Donahue et~al.(2017)Donahue, Hendricks, Rohrbach, Venugopalan,
  Guadarrama, Saenko and Darrell}]{Donahue2017PAMI}
Donahue J, Hendricks LA, Rohrbach M, Venugopalan S, Guadarrama S, Saenko K and
  Darrell T (2017) Long-term recurrent convolutional networks for visual
  recognition and description.
\newblock \emph{IEEE Trans. Pattern Anal. Mach. Intell.} 39(4): 677--691.
\newblock \doi{10.1109/TPAMI.2016.2599174}.
\newblock \urlprefix\url{https://doi.org/10.1109/TPAMI.2016.2599174}.

\bibitem[{Ferrucci et~al.(2010)Ferrucci, Brown, Chu-Carroll, Fan, Gondek,
  Kalyanpur, Lally, Murdock, Nyberg, Prager, Schlaefer and
  Welty}]{FerrucciEtAl10aimag}
Ferrucci D, Brown E, Chu-Carroll J, Fan J, Gondek D, Kalyanpur AA, Lally A,
  Murdock JW, Nyberg E, Prager J, Schlaefer N and Welty C (2010) Building
  {Watson}: An overview of the {DeepQA} project.
\newblock \emph{AI Magazine} 31(3): 59--79.
\newblock
  \urlprefix\url{http://www.aaai.org/ojs/index.php/aimagazine/article/view/2303}.

\bibitem[{Fiorini and Abel(2010)}]{fiorini_review_2010}
Fiorini SR and Abel M (2010) A review on knowledge-based computer vision .

\bibitem[{Girshick et~al.(2014)Girshick, Donahue, Darrell and
  Malik}]{girshick14CVPR}
Girshick R, Donahue J, Darrell T and Malik J (2014) Rich feature hierarchies
  for accurate object detection and semantic segmentation.
\newblock In: \emph{Computer Vision and Pattern Recognition}.

\bibitem[{Goron et~al.(2012)Goron, Marton, Lazea and Beetz}]{goron12robotik}
Goron LC, Marton ZC, Lazea G and Beetz M (2012) Segmenting cylindrical and
  box-like objects in cluttered {3D} scenes.
\newblock In: \emph{7th German Conference on Robotics (ROBOTIK)}. Munich,
  Germany.

\bibitem[{Hager(2014)}]{hager2014keynote}
Hager GD (2014) Life in a world of ubiquitous sensing.
\newblock In: \emph{Conference Keynote at IROS}.
\newblock
  \urlprefix\url{http://www.cs.jhu.edu/~hager/Talks/IROS-2014-Sensing-Keynote.pdf}.

\bibitem[{Hartz and Neumann(2007)}]{hartz_learning_2007}
Hartz J and Neumann B (2007) Learning a knowledge base of ontological concepts
  for high-level scene interpretation.
\newblock In: \emph{Machine Learning and Applications, 2007. {ICMLA} 2007.
  Sixth International Conference on}. {IEEE}, pp. 436--443.
\newblock
  \urlprefix\url{http://ieeexplore.ieee.org/xpls/abs_all.jsp?arnumber=4457269}.

\bibitem[{Hinterstoisser et~al.(2011)Hinterstoisser, Cagniart, Ilic, Konolige,
  Navab and Lepetit}]{hinterstoisser2011linemod}
Hinterstoisser S S~Holzer, Cagniart C, Ilic S, Konolige K, Navab N and Lepetit
  V (2011) Multimodal templates for real-time detection of texture-less objects
  in heavily cluttered scenes.
\newblock In: \emph{IEEE International Conference on Computer Vision (ICCV)}.

\bibitem[{Huang et~al.(2016)Huang, Rathod, Sun, Zhu, Korattikara, Fathi,
  Fischer, Wojna, Song, Guadarrama and Murphy}]{convDetectorsSurvey}
Huang J, Rathod V, Sun C, Zhu M, Korattikara A, Fathi A, Fischer I, Wojna Z,
  Song Y, Guadarrama S and Murphy K (2016) Speed/accuracy trade-offs for modern
  convolutional object detectors.
\newblock \emph{CoRR} abs/1611.10012.
\newblock \urlprefix\url{http://arxiv.org/abs/1611.10012}.

\bibitem[{Hudelot(2005)}]{hudelot_towards_2005}
Hudelot C (2005) \emph{Towards a Cognitive Vision Platform for Semantic Image
  Interpretation; Application to the Recognition of Biological Organisms}.
\newblock PhD Thesis, Université Nice Sophia Antipolis.

\bibitem[{Izadi et~al.(2011)Izadi, Kim, Hilliges, Molyneaux, Newcombe, Kohli,
  Shotton, Hodges, Freeman, Davison and Fitzgibbon}]{Izadi11KinectFusion}
Izadi S, Kim D, Hilliges O, Molyneaux D, Newcombe R, Kohli P, Shotton J, Hodges
  S, Freeman D, Davison A and Fitzgibbon A (2011) Kinectfusion: real-time 3d
  reconstruction and interaction using a moving depth camera.
\newblock In: \emph{Proceedings of the 24th annual ACM symposium on User
  interface software and technology}, UIST '11. New York, NY, USA: ACM.
\newblock ISBN 978-1-4503-0716-1, pp. 559--568.
\newblock \doi{10.1145/2047196.2047270}.
\newblock \urlprefix\url{http://doi.acm.org/10.1145/2047196.2047270}.

\bibitem[{Jia et~al.(2014)Jia, Shelhamer, Donahue, Karayev, Long, Girshick,
  Guadarrama and Darrell}]{jia2014caffe}
Jia Y, Shelhamer E, Donahue J, Karayev S, Long J, Girshick R, Guadarrama S and
  Darrell T (2014) Caffe: Convolutional architecture for fast feature
  embedding.
\newblock \emph{arXiv preprint arXiv:1408.5093} .

\bibitem[{Li et~al.(2018)Li, Bai and Hager}]{DBLP:journals/corr/abs-1803-08103}
Li C, Bai J and Hager GD (2018) A unified framework for multi-view multi-class
  object pose estimation.
\newblock \emph{CoRR} abs/1803.08103.
\newblock \urlprefix\url{http://arxiv.org/abs/1803.08103}.

\bibitem[{Lin et~al.(2017)Lin, Goyal, Girshick, He and
  Doll{\'a}r}]{lin2017focal}
Lin TY, Goyal P, Girshick R, He K and Doll{\'a}r P (2017) Focal loss for dense
  object detection.
\newblock \emph{arXiv preprint arXiv:1708.02002} .

\bibitem[{Lisca et~al.(2015)Lisca, Nyga, B\'alint-Bencz\'edi, Langer and
  Beetz}]{lisca15osd}
Lisca G, Nyga D, B\'alint-Bencz\'edi F, Langer H and Beetz M (2015) {Towards
  Robots Conducting Chemical Experiments}.
\newblock In: \emph{IEEE/RSJ International Conference on Intelligent Robots and
  Systems (IROS)}. Hamburg, Germany.
\newblock
  \urlprefix\url{http://ieeexplore.ieee.org/stamp/stamp.jsp?tp=&arnumber=7354110}.

\bibitem[{Liu et~al.(2016)Liu, Anguelov, Erhan, Szegedy, Reed, Fu and
  Berg}]{liu2016ssd}
Liu W, Anguelov D, Erhan D, Szegedy C, Reed S, Fu CY and Berg AC (2016) Ssd:
  Single shot multibox detector.
\newblock In: \emph{European Conference on Computer Vision}. Springer, pp.
  21--37.

\bibitem[{Lysenkov et~al.(2012)Lysenkov, Eruhimov and
  Bradski}]{Lysenkov-RSS-12}
Lysenkov I, Eruhimov V and Bradski G (2012) {Recognition and Pose Estimation of
  Rigid Transparent Objects with a Kinect Sensor}.
\newblock In: \emph{Proceedings of Robotics: Science and Systems}. Sydney,
  Australia.

\bibitem[{M{\"o}rwald et~al.(2010)M{\"o}rwald, Prankl, Richtsfeld, Zillich and
  Vincze}]{morwald2010}
M{\"o}rwald T, Prankl J, Richtsfeld A, Zillich M and Vincze M (2010) Blort- the
  blocks world robotic vision toolbox.
\newblock In: \emph{''Best Practice Algorithms in 3D Perception and Modeling
  for Mobile Manipulation Workshop'' - CD (in conjunction with the IEEE ICRA
  2010)}.

\bibitem[{Mozos et~al.(2011)Mozos, Marton and Beetz}]{mozos11furniture}
Mozos OM, Marton ZC and Beetz M (2011) {Furniture Models Learned from the WWW
  -- Using Web Catalogs to Locate and Categorize Unknown Furniture Pieces in 3D
  Laser Scans}.
\newblock \emph{Robotics \& Automation Magazine} 18(2): 22--32.

\bibitem[{Muja et~al.(2011)Muja, Rusu, Bradski and Lowe}]{rein11ICRA}
Muja M, Rusu RB, Bradski G and Lowe D (2011) Rein - a fast, robust, scalable
  recognition infrastructure.
\newblock In: \emph{ICRA}. Shanghai, China.

\bibitem[{Nyga et~al.(2014)Nyga, Balint-Benczedi and Beetz}]{icra14ensmln}
Nyga D, Balint-Benczedi F and Beetz M (2014) {PR2 Looking at Things: Ensemble
  Learning for Unstructured Information Processing with Markov Logic Networks}.
\newblock In: \emph{IEEE International Conference on Robotics and Automation
  (ICRA)}. Hong Kong, China.
\newblock
  \urlprefix\url{http://ieeexplore.ieee.org/stamp/stamp.jsp?tp=\&arnumber=6907427}.

\bibitem[{Okada et~al.(2007)Okada, Kojima, Tokutsu, Maki, Mori and
  Inaba}]{okada2007iros}
Okada K, Kojima M, Tokutsu S, Maki T, Mori Y and Inaba M (2007) {Multi-cue 3D
  object recognition in knowledge-based vision-guided humanoid robot system}.
\newblock \emph{IEEE/RSJ International Conference on Intelligent Robots and
  Systems (IROS).} : 3217--3222.

\bibitem[{OpenCyc()}]{opencyc}
OpenCyc (2009) {OpenCyc}.
\newblock {\tt www.opencyc.org }.

\bibitem[{Pangercic et~al.(2010)Pangercic, Tenorth, Jain and
  Beetz}]{iros10kcopman}
Pangercic D, Tenorth M, Jain D and Beetz M (2010) {Combining Perception and
  Knowledge Processing for Everyday Manipulation}.
\newblock In: \emph{IEEE/RSJ International Conference on Intelligent Robots and
  Systems (IROS)}. Taipei, Taiwan, pp. 1065--1071.

\bibitem[{Pangercic et~al.(2012)Pangercic, Tenorth, Pitzer and
  Beetz}]{iros12semantic_mapping}
Pangercic D, Tenorth M, Pitzer B and Beetz M (2012) Semantic object maps for
  robotic housework - representation, acquisition and use.
\newblock In: \emph{{2012 IEEE/RSJ International Conference on Intelligent
  Robots and Systems (IROS)}}. Vilamoura, Portugal.

\bibitem[{Pauwels and Kragic(2015)}]{pauwels_simtrack_2015}
Pauwels K and Kragic D (2015) Simtrack: A simulation-based framework for
  scalable real-time object pose detection and tracking.
\newblock In: \emph{IEEE/RSJ International Conference on Intelligent Robots and
  Systems}. Hamburg, Germany.

\bibitem[{Pech-Pacheco et~al.(2010)Pech-Pacheco, Chamorro-Martinez and
  Fernandez-Valdivia}]{blur@icpr2000}
Pech-Pacheco J, Chamorro-Martinez J and Fernandez-Valdivia J (2010) Diatom
  autofocusing in bright eld microscopy: a comparative study.
\newblock \emph{International Conference on Pattern Recognition} .

\bibitem[{Pronobis et~al.(2010)Pronobis, Mozos, Caputo and
  Jensfelt}]{Pronobis10IJRR}
Pronobis A, Mozos OM, Caputo B and Jensfelt P (2010) Multi-modal semantic place
  classification.
\newblock \emph{The International Journal of Robotics Research (IJRR), Special
  Issue on Robotic Vision} 29(2-3): 298--320.
\newblock \doi{10.1177/0278364909356483}.
\newblock
  \urlprefix\url{http://www.pronobis.pro/publications/pronobis2010ijrr}.

\bibitem[{Redmon and Farhadi(2016)}]{redmon2016yolo9000}
Redmon J and Farhadi A (2016) Yolo9000: Better, faster, stronger.
\newblock \emph{arXiv preprint arXiv:1612.08242} .

\bibitem[{Rudolph(2011)}]{Rudolph2011}
Rudolph S (2011) \emph{Foundations of Description Logics}.
\newblock Berlin, Heidelberg: Springer Berlin Heidelberg.
\newblock ISBN 978-3-642-23032-5, pp. 76--136.
\newblock \doi{10.1007/978-3-642-23032-5_2}.
\newblock \urlprefix\url{https://doi.org/10.1007/978-3-642-23032-5_2}.

\bibitem[{Russell and Norvig(2010)}]{russell10aima}
Russell SJ and Norvig P (2010) \emph{Artificial Intelligence --- A Modern
  Approach (3. internat. ed.)}.
\newblock Pearson Education.

\bibitem[{Rusu and Cousins(2011)}]{Rusu_ICRA2011_PCL}
Rusu RB and Cousins S (2011) {3D is here: Point Cloud Library (PCL)}.
\newblock In: \emph{{IEEE International Conference on Robotics and Automation
  (ICRA)}}. Shanghai, China, pp. 1--4.

\bibitem[{Scheutz(2006)}]{scheutz06aai}
Scheutz M (2006) {ADE} - steps towards a distributed development and runtime
  environment for complex robotic agent architectures.
\newblock \emph{Applied Artificial Intelligence} 20(4-5).

\bibitem[{Sko\v{c}aj et~al.(2016)Sko\v{c}aj, Vre\v{c}ko, Mahni\v{c},
  Jan\'{i}\v{c}ek, Kruijff, Hanheide, Hawes, Wyatt, Keller, Zhou, Zillich and
  Kristan}]{skocajInteractive}
Sko\v{c}aj D, Vre\v{c}ko A, Mahni\v{c} M, Jan\'{i}\v{c}ek M, Kruijff GJM,
  Hanheide M, Hawes N, Wyatt JL, Keller T, Zhou K, Zillich M and Kristan M
  (2016) An integrated system for interactive continuous learning of
  categorical knowledge.
\newblock \emph{Journal of Experimental \& Theoretical Artificial Intelligence}
  28(5): 823--848.
\newblock \doi{10.1080/0952813X.2015.1132268}.
\newblock \urlprefix\url{https://doi.org/10.1080/0952813X.2015.1132268}.

\bibitem[{Socher et~al.(2014)Socher, Karpathy, Le, Manning and
  Ng}]{Socher2014GroundedCS}
Socher R, Karpathy A, Le QV, Manning CD and Ng AY (2014) Grounded compositional
  semantics for finding and describing images with sentences.
\newblock \emph{TACL} 2: 207--218.

\bibitem[{Sridharan et~al.(2010)Sridharan, Wyatt and
  Dearden}]{Sridharan10planningtToSee}
Sridharan M, Wyatt J and Dearden R (2010) Planning to see: A hierarchical
  approach to planning visual actions on a robot using pomdps.
\newblock \emph{Artificial Intelligence} 174(11): 704 -- 725.
\newblock \doi{https://doi.org/10.1016/j.artint.2010.04.022}.
\newblock
  \urlprefix\url{http://www.sciencedirect.com/science/article/pii/S0004370210000664}.

\bibitem[{Tenorth(2011)}]{tenorth2011phd}
Tenorth M (2011) \emph{Knowledge Processing for Autonomous Robots}.
\newblock PhD Thesis, Technische Universit\"at M\"unchen.
\newblock
  \urlprefix\url{http://nbn-resolving.de/urn/resolver.pl?urn:nbn:de:bvb:91-diss-20111125-1079930-1-7}.

\bibitem[{Tenorth and Beetz(2013)}]{tenorth13knowrob}
Tenorth M and Beetz M (2013) {KnowRob -- A Knowledge Processing Infrastructure
  for Cognition-enabled Robots}.
\newblock \emph{Int. Journal of Robotics Research} 32(5): 566 -- 590.
\newblock \urlprefix\url{http://ijr.sagepub.com/content/32/5/566.short}.

\bibitem[{Thrun et~al.(2005)Thrun, Burgard and Fox}]{thrun05probabilistic}
Thrun S, Burgard W and Fox D (2005) \emph{Probabilistic Robotics}.
\newblock Cambridge: MIT Press.

\bibitem[{Tombari and Stefano(2010)}]{tombari10clutter}
Tombari F and Stefano LD (2010) Object recognition in 3d scenes with occlusions
  and clutter by hough voting.
\newblock \emph{Proceedings of the 4th Pacific-Rim Symposium on Image and Video
  Technology (PSIVT 2010)} 0: 349--355.

\bibitem[{Town(2006)}]{town_ontological_2006}
Town C (2006) Ontological inference for image and video analysis.
\newblock \emph{Machine Vision and Applications} 17(2): 94--115.
\newblock
  \urlprefix\url{http://link.springer.com/article/10.1007/s00138-006-0017-3}.

\bibitem[{Wiedemeyer et~al.(2015)Wiedemeyer, Balint-Benczedi and
  Beetz}]{Wiedemeyer15pervasive}
Wiedemeyer T, Balint-Benczedi F and Beetz M (2015) Pervasive 'calm' perception
  for autonomous robotic agents.
\newblock In: \emph{Proceedings of the 2015 International Conference on
  Autonomous Agents and Multiagen Systems}. Istanbul, Turkey: ACM.
\newblock \urlprefix\url{http://dl.acm.org/citation.cfm?id=2773264}.

\bibitem[{Worch et~al.(2015)Worch, B{\'a}lint-Bencz{\'e}di and
  Beetz}]{worch2015perceptionHRI}
Worch JH, B{\'a}lint-Bencz{\'e}di F and Beetz M (2015) Perception for everyday
  human robot interaction.
\newblock \emph{KI - K{\"u}nstliche Intelligenz} 30(1): 21--27.
\newblock \doi{10.1007/s13218-015-0400-1}.
\newblock \urlprefix\url{http://dx.doi.org/10.1007/s13218-015-0400-1}.

\bibitem[{Wyatt et~al.(2010)Wyatt, Aydemir, Brenner, Hanheide, Hawes, Jensfelt,
  Kristan, Kruijff, Lison, Pronobis, Sj\"{o}\"{o}, Sko\v{c}aj, Vre\v{c}ko,
  Zender and Zillich}]{wyatt10tamd}
Wyatt JL, Aydemir A, Brenner M, Hanheide M, Hawes N, Jensfelt P, Kristan M,
  Kruijff GJM, Lison P, Pronobis A, Sj\"{o}\"{o} K, Sko\v{c}aj D, Vre\v{c}ko A,
  Zender H and Zillich M (2010) Self-understanding and self-extension: A
  systems and representational approach.
\newblock \emph{IEEE Transactions on Autonomous Mental Development} 2(4): 282
  -- 303.

\end{thebibliography}
\end{document}